\documentclass[10pt,twocolumn,letterpaper]{article}

\usepackage{cvpr}              

\usepackage{graphicx}
\usepackage{amsmath}
\usepackage{amssymb}
\usepackage{booktabs}

\usepackage[pagebackref,breaklinks,colorlinks]{hyperref}

\usepackage[capitalize]{cleveref}
\crefname{section}{Sec.}{Secs.}
\Crefname{section}{Section}{Sections}
\Crefname{table}{Table}{Tables}
\crefname{table}{Tab.}{Tabs.}

\usepackage{algorithm,algorithmicx,algpseudocode}
\usepackage{float}

\usepackage{multirow, booktabs}
\usepackage{siunitx}
\usepackage{makecell} 

\usepackage{booktabs}       
\usepackage{mathtools,amssymb}
\usepackage{amsfonts}       
\usepackage{nicefrac}       
\usepackage{microtype}      
\usepackage{array,colortbl}
\usepackage{xcolor}
\usepackage{algorithm,algorithmicx,algpseudocode}
\usepackage{graphbox}
\usepackage{placeins}
\usepackage{wrapfig}
\usepackage{subcaption}
\usepackage{etoolbox}

\newlength\myindent
\newcommand{\defeq}{\coloneqq}
\newcommand{\grad}{\nabla}
\newcommand{\E}{\mathbb{E}}

\newcommand{\Eb}[2]{\E_{#1}\!\left[#2\right]}

\newcommand{\bI}{\mathbf{I}}

\newcommand{\bzero}{\mathbf{0}}

\newcommand{\bx}{\mathbf{x}}

\newcommand{\bz}{\mathbf{z}}

\newcommand{\bepsilon}{{\boldsymbol{\epsilon}}}
\newcommand{\bmu}{{\boldsymbol{\mu}}}

\newcommand{\bSigma}{{\boldsymbol{\Sigma}}}

\usepackage{marvosym, ifsym}

\def\confName{CVPR}
\def\confYear{2023}

\begin{document}

\title{Collaborative Diffusion for Multi-Modal Face Generation and Editing}


\author{Ziqi Huang
\quad
Kelvin C.K. Chan
\quad
Yuming Jiang
\quad
Ziwei Liu\textsuperscript{\Letter}\\
S-Lab, Nanyang Technological University
\quad
\\
{\tt\small \{ziqi002, chan0899, yuming002, ziwei.liu\}@ntu.edu.sg}\\
}


\twocolumn[{%
            \renewcommand\twocolumn[1][]{#1}%
            \vspace{-3em}
            \maketitle
            \begin{center}
                \centering
                \includegraphics[width=0.99\textwidth]{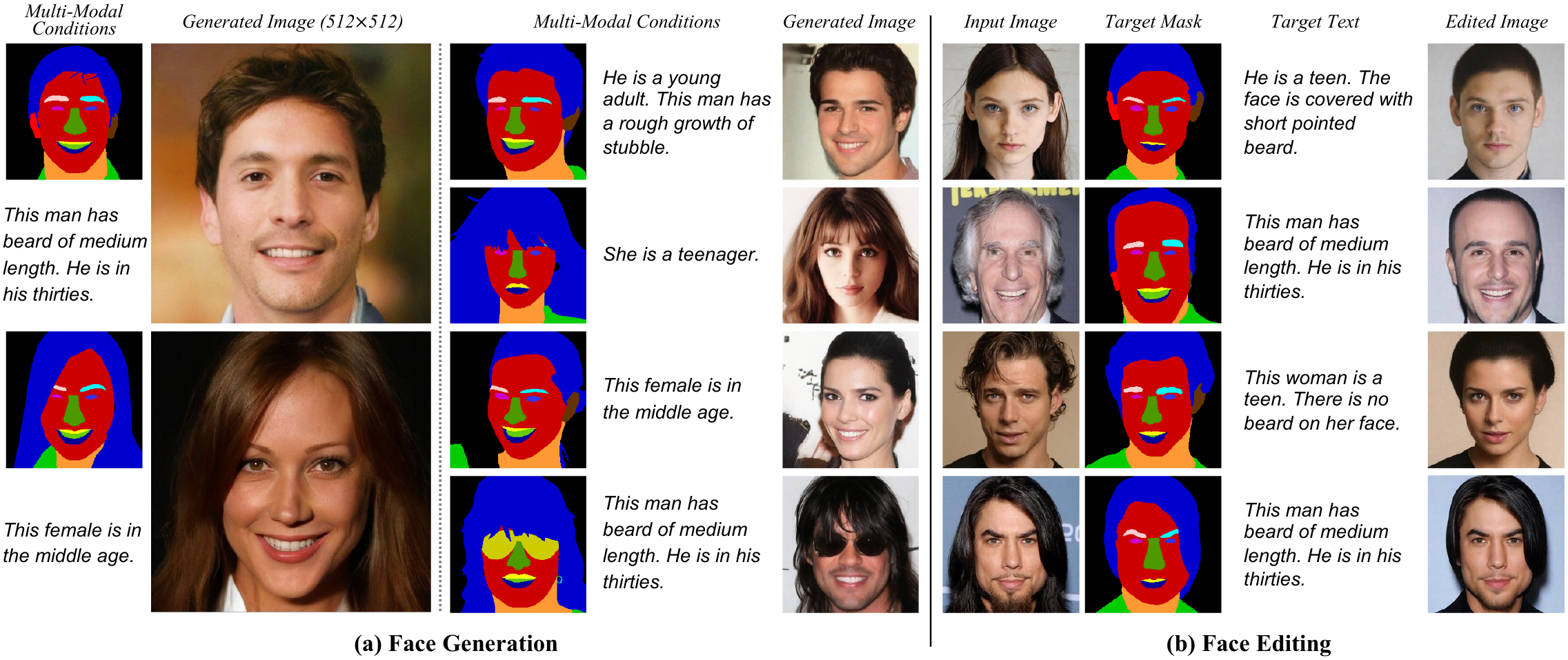}
                \captionof{figure}{
                We propose \textbf{\textit{Collaborative Diffusion}}, where users can use multiple modalities to control face generation and editing. 
                \textbf{(a) Face Generation}. Given multi-modal controls, our framework synthesizes high-quality images consistent with the input conditions. 
                \textbf{(b) Face Editing}. Collaborative Diffusion  also supports multi-modal editing of real images with promising identity preservation capability.
                }
                \label{teaser}
                \vspace{5pt}
            \end{center}%
        }]


\begin{abstract}

Diffusion models arise as a powerful generative tool recently. 
Despite the great progress, existing diffusion models mainly focus on uni-modal control, \ie, the diffusion process is driven by only one modality of condition.
To further unleash the users' creativity, it is desirable for the model to be controllable by multiple modalities simultaneously, \eg generating and editing faces by describing the age (text-driven) while drawing the face shape (mask-driven).
\makeatletter{\renewcommand*{\@makefnmark}{}
\footnotetext{\hspace{-10pt}\scriptsize\textsuperscript{\Letter}Corresponding author.\makeatother}
\footnotetext{\hspace{-10pt}\scriptsize Project page: \href{https://ziqihuangg.github.io/projects/collaborative-diffusion.html}{https://ziqihuangg.github.io/projects/collaborative-diffusion.html}\makeatother}
\footnotetext{\hspace{-10pt}\scriptsize Code: \href{https://github.com/ziqihuangg/Collaborative-Diffusion}{https://github.com/ziqihuangg/Collaborative-Diffusion}}\makeatother}

In this work, we present \textbf{Collaborative Diffusion}, where pre-trained uni-modal diffusion models collaborate to achieve multi-modal face generation and editing without re-training.
Our key insight is that diffusion models driven by different modalities are inherently complementary regarding the latent denoising steps, where bilateral connections can be established upon.
Specifically, we propose dynamic diffuser, a meta-network that adaptively hallucinates multi-modal denoising steps by predicting the spatial-temporal influence functions for each pre-trained uni-modal model.
Collaborative Diffusion not only collaborates generation capabilities from uni-modal diffusion models, but also integrates multiple uni-modal manipulations to perform multi-modal editing.
Extensive qualitative and quantitative experiments demonstrate the superiority of our framework in both image quality and condition consistency.

\end{abstract}
\section{Introduction}

Recent years have witnessed substantial progress in image synthesis and editing with the surge of diffusion models~\cite{ho2020ddpm, sohl2015deep, song2020score, dhariwal2021beatgan}. In addition to the remarkable synthesis quality, one appealing property of diffusion models is the flexibility of conditioning on various modalities, such as texts~\cite{nichol2021glide, rombach2022ldm, gu2022vqdiffusion, kim2022diffusionclip, avrahami2022blended, saharia2022imagen, kawar2022imagic}, segmentation masks~\cite{rombach2022ldm, wang2022semantic, wang2022piti}, and sketches~\cite{cheng2023adaptively, wang2022piti}. However, existing explorations are largely confined to the use of a single modality at a time. The exploitation of multiple conditions remains under-explored. As a generative tool, its controllability is still limited.

To unleash users' creativity, it is desirable that the model is simultaneously controllable by multiple modalities. 
While it is trivial to extend the current supervised framework with multiple modalities, training a large-scale model from scratch is computationally expensive, especially when extensive hyper-parameter tuning and delicate architecture designs are needed.
More importantly, each trained model could only accept a fixed combination of modalities, and hence re-training is necessary when a subset of modalities are absent, or when additional modalities become available.
The above demonstrates the necessity of a unified framework that effectively exploits pre-trained models and integrate them for multi-modal synthesis and editing. 

In this paper, we propose \textbf{\textit{Collaborative Diffusion}}, a framework that synergizes pre-trained uni-modal diffusion models for multi-modal face generation and editing without the need of re-training. 
Motivated by the fact that different modalities are complementary to each other~(\eg, \textit{text} for age and \textit{mask} for hair shape), we explore the possibility of establishing lateral connections between models driven by different modalities. We propose \textit{dynamic diffuser} to adaptively predict the spatial-temporal \textit{influence function} for each pre-trained model. The \textit{dynamic diffuser} dynamically determines spatial-varying and temporal-varying influences of each model, suppressing contributions from irrelevant modalities while enhancing contributions from admissible modalities.
In addition to multi-modal synthesis, the simplicity and flexibility of our framework enable extension to multi-modal face editing with minimal modifications. In particular, the \textit{dynamic diffuser} is first trained for collaborative synthesis. It is then fixed and combined with existing face editing approaches~\cite{kawar2022imagic, ruiz2022dreambooth} for multi-modal editing. It is worth-mentioning that users can select the best editing approaches based on their needs without the need of altering the \textit{dynamic diffusers}.

We demonstrate both qualitatively and quantitatively that our method achieves superior image quality and condition consistency in both synthesis and editing tasks.
Our contributions can be summarized as follows:

\vspace{-0.4em}
\begin{itemize}
    \setlength\itemsep{0em}
    \item We introduce \textbf{\textit{Collaborative Diffusion}}, which exploits pre-trained uni-modal diffusion models for multi-modal controls without re-training. Our approach is the first attempt towards flexible integration of uni-modal diffusion models into a single collaborative framework.
    \item Tailored for the iterative property of diffusion models, we propose \textit{dynamic diffuser}, which predicts the spatial-varying and temporal-varying \textit{influence functions} to selectively enhance or suppress the contributions of the given modalities at each iterative step.
    \item We demonstrate the flexibility of our framework by extending it to face editing driven by multiple modalities. Both quantitative and qualitative results demonstrate the superiority of \textit{Collaborative Diffusion} in multi-modal face generation and editing.
\end{itemize}

\section{Related Work}
\noindent\textbf{Diffusion Models.}
Diffusion models~\cite{ho2020ddpm, sohl2015deep, song2020score} have recently become a mainstream approach for image synthesis~\cite{dhariwal2021beatgan, esser2021imagebart, meng2021sdedit} apart from Generative Adversarial Networks (GANs)~\cite{goodfellow2014gan}, and success has also been found in various domains including video generation~\cite{harvey2022fdm,villegas2022phenaki,singer2022makeavideo,ho2022imagenvideo}, image restoration~\cite{saharia2022sr3, ho2022cascaded}, semantic segmentation~\cite{baranchuk2021label,graikos2022diffusion, amit2021segdiff}, and natural language processing~\cite{austin2021structured}.
In the diffusion-based framework, models are trained with score-matching objectives~\cite{hyvarinen2005estimation, vincent2011connection} at various noise levels, and sampling is done via iterative denoising. 
Existing works focus on improving the performance and efficiency of diffusion models through enhanced architecture designs~\cite{rombach2022ldm, gu2022vqdiffusion} and sampling schemes~\cite{song2020ddim}. In contrast, this work focuses on exploiting existing models, and providing a succinct framework for multi-modal synthesis and editing without large-scale re-training of models.

\noindent\textbf{Face Generation.}
Existing face generation approaches can be divided into three main directions.
Following the GAN paradigm, the \textit{StyleGAN} series~\cite{Karras2019stylegan1, karras2020stylegan2, Karras2021stylegan3} boost the quality of facial synthesis, and provide an interpretable latent space for steerable style controls and manipulations.
The \textit{vector-quantized} approaches~\cite{van2017vqvae, esser2021vqgan} learn a discrete codebook by mapping the input images into a low-dimensional discrete feature space. The learned codebook is then sampled, either sequentially~\cite{van2017vqvae, esser2021vqgan} or parallelly~\cite{chang2022maskgit, bond2022unleashing, gu2022vqdiffusion}, for synthesis. 
In contrast to the previous two approaches, \textit{diffusion models} are trained with a stationary objective, without the need of optimizing complex losses (\eg, adversarial loss) or balancing multiple objectives (\eg, codebook loss versus reconstruction loss). With training simplicity as a merit, diffusion-based approaches have become increasingly popular in recent years. Our framework falls in the diffusion-based paradigm. In particular, we leverage pre-trained diffusion models for multi-modal generation and editing.

\begin{figure*}[t]
  \centering
   \includegraphics[width=1.0\linewidth]{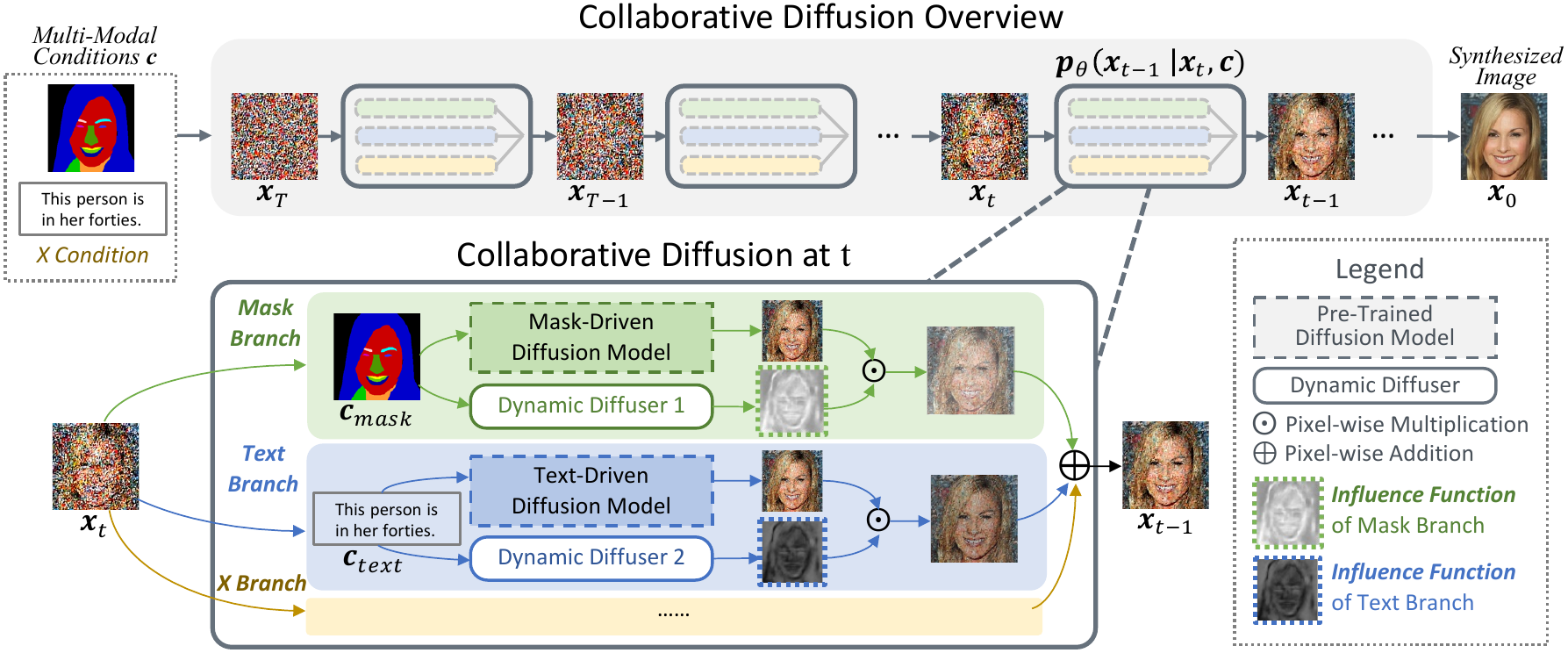}
   \caption{\textbf{Overview of Collaborative Diffusion}. 
   We use pre-trained uni-modal diffusion models to perform multi-modal guided face generation and editing. 
   At each step of the reverse process (\ie, from timestep $t$ to $t-1$), the \textit{dynamic diffuser} predicts the spatial-varying and temporal-varying \textit{influence function} to selectively enhance or suppress the contributions of the given modality.
   }
   \label{fig:framework}
   \vspace{10pt}
\end{figure*}

\noindent\textbf{Conditional Face Generation and Editing.}
Conditional generation~\cite{xia2021tedigan1, xia2021tedigan2, park2019spade, ding2021cogview, ramesh2022dalle2, reed2016generative, reed2016learning, zhang2017stackgan, zhang2018stackgan++, xu2018attngan, li2019object, koh2021text, esser2021imagebart, liu2023more, saharia2022imagen, ramesh2021dalle1, gafni2022makeascene, wang2022piti} and editing~\cite{xia2021tedigan1, xia2021tedigan2, patashnik2021styleclip, lee2020maskgan, shen2020interfacegan1, shen2020interfacegan2, couairon2022diffedit} is an active line of research focusing on conditioning generative models on different modalities, such as texts~\cite{xia2021tedigan1, xia2021tedigan2, patashnik2021styleclip, jiang2022text2human}, segmentation masks~\cite{lee2020maskgan, richardson2021encoding, park2019spade, li2020manigan}, and audios~\cite{song2018talking}. For example, StyleCLIP~\cite{patashnik2021styleclip}, DiffusionCLIP~\cite{kim2022diffusionclip}, and many others~\cite{sun2022anyface, li2022stylet2i} have demonstrated remarkable performance in text-guided face generation and editing.
However, most existing models do not support simultaneous conditioning on multiple modalities (\eg, text and mask at the same time), and supporting additional modalities often requires time-consuming model re-training and extensive hyper-parameter tuning, which are not preferable in general. 
In this work, we propose \textit{Collaborative Diffusion} to exploit pre-trained uni-modal diffusion models~\cite{rombach2022ldm} (\eg, text-driven and mask-driven models) to achieve multi-modal conditioning without model re-training.

\section{Collaborative Diffusion}

We propose \textbf{\textit{Collaborative Diffusion}}, which exploits multiple pre-trained uni-modal diffusion models (Section~\ref{subsec:unimodalDM}) for multi-modal generation and editing. 
The key of our framework is the \textit{dynamic diffuser}, which adaptively predicts the \textit{influence functions} to enhance or suppress the contributions of the pre-trained models based on the spatial-temporal influences of the modalities. 
Our framework is compatible with most existing approaches for both multi-modal guided synthesis (Section~\ref{subsec:collaborativesynthesis}) and multi-modal editing (Section~\ref{subsec:collaborativeediting}).

\subsection{Uni-Modal Conditional Diffusion Models} 
\label{subsec:unimodalDM}
Diffusion models are a class of generative models that model the data distribution in the form of $p_\theta(\bx_0) \defeq \int p_\theta(\bx_{0:T}) \,d\bx_{1:T}$. 
%
The \textit{diffusion process} (\textit{a.k.a.} \textit{forward process}) gradually adds Gaussian noise to the data and eventually corrupts the data $\bx_0$ into an approximately pure Gaussian noise $\bx_T$ using a variance schedule $\beta_1, \dotsc, \beta_T$: 
\begin{gather}
\begin{split}
    &q(\bx_{1:T} | \bx_0) \defeq \prod_{t=1}^T q(\bx_t | \bx_{t-1} ), \qquad \\
    &q(\bx_t|\bx_{t-1}) \defeq \mathcal{N}(\bx_t;\sqrt{1-\beta_t}\bx_{t-1},\beta_t \bI). \label{eq:forwardprocess}
\end{split}
\end{gather}
%
Reversing the \textit{forward process} allows sampling new data $\bx_0$ by starting from $p(\bx_T)=\mathcal{N}(\bx_T; \bzero, \bI)$. The \textit{reverse process} is defined as a Markov chain where each step is a learned Gaussian transition $(\bmu_\theta, \bSigma_\theta)$:
\begin{gather}
\begin{split}
  &p_\theta(\bx_{0:T}) \defeq p(\bx_T)\prod_{t=1}^T p_\theta(\bx_{t-1}|\bx_t), \qquad \\
  &p_\theta(\bx_{t-1}|\bx_t) \defeq \mathcal{N}(\bx_{t-1}; \bmu_\theta(\bx_t, t), \bSigma_\theta(\bx_t, t)).
\end{split}
\end{gather}
Training diffusion models relies on minimizing the variational bound on $p(x)$'s negative log-likelihood. The commonly used optimization objective $L_\mathrm{DM}$~\cite{ho2020ddpm} reparameterizes the learnable Gaussian transition as $\bepsilon_\theta(\cdot)$, and temporally reweights the variational bound to trade for better sample quality:
\begin{gather}
 L_\mathrm{DM}(\theta) \defeq \Eb{t, \bx_0, \bepsilon \sim \mathcal{N}(\bzero, \bI)}{ \left\| \bepsilon - \bepsilon_\theta(\bx_t, t) \right\|^2}, \label{eq:training_objective_simple}
\end{gather}
where $\bx_t$ can be directly approximated by $\bx_t = \sqrt{\bar\alpha_t} \bx_0 + \sqrt{1-\bar\alpha_t}\bepsilon$, with $\bar\alpha_t \defeq \prod_{s=1}^t \alpha_s$ and $\alpha_t \defeq 1-\beta_t$.

To sample data $\bx_0$ from a trained diffusion model $\bepsilon_\theta(\cdot)$, we iteratively denoise $\bx_t$ from $t = T$ to $t = 1$ with noise $\textbf{z}$:  
\begin{gather}
    \bx_{t-1} = \frac{1}{\sqrt{\alpha_t}}\left( \bx_t - \frac{1-\alpha_t}{\sqrt{1-\bar\alpha_t}} \bepsilon_\theta(\bx_t, t) \right) + \sigma_t \bz
\end{gather}

The unconditional diffusion models can be extended to model conditional distributions $p_\theta(\bx_0 | c)$, where $\bx_0$ is the image corresponding to the condition $c$ such as class labels, segmentation masks, and text descriptions. The conditional diffusion model receives an additional input $\tau(c)$ and is trained by minimizing ${ \left\| \bepsilon - \bepsilon_\theta(\bx_t, t, \tau(c)) \right\|^2}$, where $\tau(\cdot)$ is an encoder that projects the condition $c$ to an embedding $\tau(c)$. For brevity, we will use $c$ to represent $\tau(c)$ in our subsequent discussions.

\algrenewcommand\algorithmicindent{0.5em}%
\begin{figure*}[t]
\begin{minipage}[t]{0.492\textwidth}
\begin{algorithm}[H]
  \caption{Dynamic Diffuser Training} \label{alg:training}
  \small
  \begin{algorithmic}[1]
    \Repeat
      \State $\bx_0, c_1, c_2, ..., c_M \sim q(\bx_0, c_1, c_2, ..., c_M)$
      \State $t \sim \mathrm{Uniform}(\{1, \dotsc, T\})$
      \State $\bepsilon\sim\mathcal{N}(\bzero,\bI)$ 
      \For{$m = 1, ..., M$} 
      \State \hskip1.0em $\bepsilon_{pred,m,t} = \bepsilon_{\theta_m}(\sqrt{\bar\alpha_t} \bx_0 + \sqrt{1-\bar\alpha_t}\bepsilon, t, c_{m})$ \vspace{0.1cm}
      \State \hskip1.0em $\textbf{I}_{m,t} = \textbf{D}_{\phi_m}(\sqrt{\bar\alpha_t} \bx_0 + \sqrt{1-\bar\alpha_t}\bepsilon, t, c_{m})$ \vspace{0.1cm}
      \EndFor
      \vspace{0.1cm}
      \State $\hat{\textbf{I}}_{m,t, p} = \frac{\exp(\textbf{I}_{m,t, p})}{\sum_{j=1}^M \exp(\textbf{I}_{j,t, p})}$, softmax at each pixel $p$ \vspace{0.1cm} 
      \State $\bepsilon_{pred,t} = \sum_{m=1}^M \hat{\textbf{I}}_{m,t} \odot \bepsilon_{pred,m,t}$  \vspace{0.1cm}
      \State Take gradient descent step on
      \Statex $\qquad \grad_\phi \left\| \bepsilon - \bepsilon_{pred,t} \right\|^2$ where $\phi = \{\phi_m | m = 1, ..., M\}$ 
    \Until{converged}
  \end{algorithmic}
\end{algorithm}
\end{minipage}
\hfill
%
\begin{minipage}[t]{0.492\textwidth}
\begin{algorithm}[H]
  \caption{Collaborative Sampling} \label{alg:sampling}
  \small
  \begin{algorithmic}[1]
    \vspace{.04in}
    \State $\bx_T \sim \mathcal{N}(\bzero, \bI)$
    \For{$t=T, \dotsc, 1$}
      \State $\bz \sim \mathcal{N}(\bzero, \bI)$ if $t > 1$, else $\bz = \bzero$  
      \For{$m = 1, ..., M$}
      \State \hskip1.0em $\bepsilon_{pred,m,t} = \bepsilon_{\theta_m}(\bx_{t}, t, c_{m})$ \vspace{0.1cm}
      \State \hskip1.0em $\textbf{I}_{m,t} = \textbf{D}_{\phi_m}(\bx_{t}, t, c_{m})$ \vspace{0.1cm}
      \EndFor \vspace{0.1cm}
      \State $\hat{\textbf{I}}_{m,t, p} = \frac{\exp(\textbf{I}_{m,t, p})}{\sum_{j=1}^M \exp(\textbf{I}_{j,t, p})}$, softmax at each pixel $p$ \vspace{0.1cm}
      \State $\bepsilon_{pred,t} = \sum_{m=1}^M \hat{\textbf{I}}_{m,t} \odot \bepsilon_{pred,m,t}$  \vspace{0.1cm}
      \State $\bx_{t-1} = \frac{1}{\sqrt{\alpha_t}}\left( \bx_t - \frac{1-\alpha_t}{\sqrt{1-\bar\alpha_t}} \bepsilon_{pred,t} \right) + \sigma_t \bz$
    \EndFor
    \State \textbf{return} $\bx_0$
    \vspace{.04in}
  \end{algorithmic}
\end{algorithm}
\end{minipage}
\vspace{-1em}
\end{figure*}
%

\subsection{Multi-Modal Collaborative Synthesis}
\label{subsec:collaborativesynthesis}
In our \textbf{\textit{Collaborative Diffusion}}, multiple uni-modal diffusion models collaborate at each step of the denoising process for multi-modal guided synthesis. The core of our framework is the \emph{dynamic diffuser}, which determines the extent of contribution from each collaborator by predicting the spatial-temporal \textit{influence functions}.

\noindent \textbf{Problem Formulation}. 
Given $M$ pre-trained uni-modal conditional diffusion models $\{\epsilon_{\theta_m}\}$ (each models the distribution $p(\bx_0 | c_m)$), where the modality index $m = 1, ..., M$, our objective is to sample from $p(\bx_0 | \textbf{c})$, where $\textbf{c} = \{c_1, c_2, \cdots, c_M\}$, without altering pre-trained models. 

\noindent \textbf{Dynamic Diffuser}. 
\begin{figure}[t]
  \centering
   \includegraphics[width=0.7\linewidth]{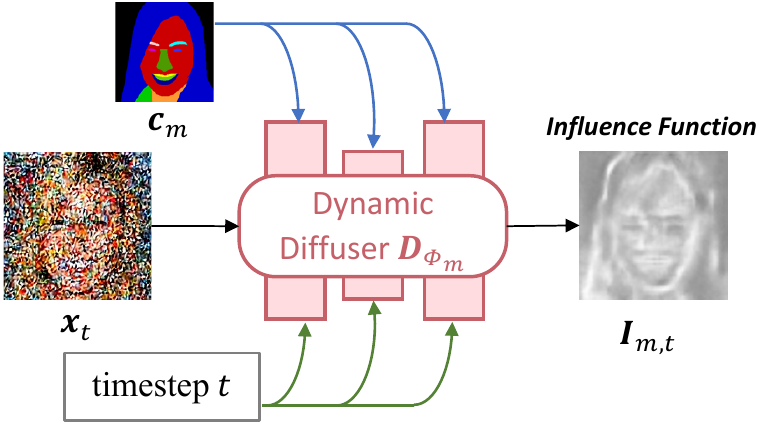}
   \caption{\textbf{Dynamic Diffuser.} It predicts the \textit{influence function} that represents the desired level of contribution from each pre-trained diffusion model. The \textit{influence function} varies at different timestep $t$ and different spatial locations, and is dependent on the state of diffusion $\bx_t$ and the condition input $c_m$.}
   \label{fig:dynamicdiffusers}
\end{figure}
In diffusion models, each step of the \emph{reverse process} requires predicting the noise $\bepsilon$. With multiple diffusion models collaborating, we need to carefully determine \textit{when}, \textit{where}, and \textit{how much} each diffusion model contributes to the prediction $\bepsilon$. At each diffusion timestep $t = T, ..., 1$, the influence $\textbf{I}_{m,t}$ from each pre-trained diffusion model $\bepsilon_{\theta_m}$ is adaptively determined by a \textit{dynamic diffuser} $\textbf{D}_{\phi_m}$: 
\begin{gather}
  \textbf{I}_{m, t} = \textbf{D}_{\phi_m}(\bx_t, t, c_{m}),
\end{gather}
where $m = 1, ..., M$ is the index of the modalities, $\textbf{I}_{m,t} \in \mathbb{R}^{h{\times}w}$, $\bx_t$ is the noisy image at time $t$, $c_m$ is the condition of the $m^{th}$ modality, and $\textbf{D}_{\phi_m}$ is the \textit{dynamic diffuser} implemented by a UNet~\cite{ronneberger2015unet}. 
To regularize the overall influence strength,
we perform softmax across all modalities' $\textbf{I}_{m,t}$ at each pixel $p$ to obtain the final \emph{influence function} $\hat{\textbf{I}}_{m,t}$:
\begin{gather}
  \hat{\textbf{I}}_{m,t, p} = \frac{\exp(\textbf{I}_{m,t, p})}{\sum_{j=1}^M \exp(\textbf{I}_{j,t, p})}.
\end{gather}
%
%

%
\noindent \textbf{Multi-Modal Collaboration}. 
We use the $M$ learned \emph{influence functions} $\hat{\textbf{I}}_{m,t}$ to control the contribution from each pre-trained diffusion model at each denoising step:
\begin{gather}
  \bepsilon_{pred,t} = \sum_{m=1}^M \hat{\textbf{I}}_{m,t} \odot \bepsilon_{\theta_m}(x_{t}, t, c_m)
\end{gather}
where $\bepsilon_{\theta_m}$ is the $m^{th}$ collaborator,  
and $\odot$ denotes pixel-wise multiplication.
%
The complete training procedure of \emph{dynamic diffusers} and the image sampling strategy in our \emph{Collaborative Diffusion} framework are detailed in Algorithm~\ref{alg:training} and Algorithm~\ref{alg:sampling}, respectively.

\noindent \textbf{Relation with Composable Diffusion}. 
Composable Diffusion~\cite{liu2022composablediffusion} combines two instances of the same text-to-image diffusion model to achieve compositional visual generation via intermediate results addition.
Our framework is related to Composable Diffusion in terms of composing diffusion models for image synthesis,  but is substantially different in terms of the task nature and methodology.
Composable Diffusion aims to decompose the text condition into elementary segments to factorize the conditional synthesis problem, while we aim to integrate uni-modal collaborators to achieve multi-modal controls.
Furthermore, different from Composable Diffusion, which can only compose instances of the same text-based synthesis model (\ie, the weights of the diffusion models being composed are the same), our framework possesses the flexibility in integrating models with different weights, architectures, and modalities through the learned \textit{dynamic diffusers}.

\subsection{Multi-Modal Collaborative Editing}
\label{subsec:collaborativeediting}
In addition to face synthesis, our framework is also capable of combining multiple facial manipulations, each of which is guided by a different modality, into a collaborative edit.
Our collaborative framework in theory can integrate any existing uni-modal diffusion-based editing approach for collaborative editing.

In this work, we demonstrate such possibility by extending Imagic~\cite{kawar2022imagic} to the multi-modal paradigm. We first follow Imagic to fine-tune the embeddings and models to better capture the input face identity during editing. The trained \emph{dynamic diffusers} discussed in Section~\ref{subsec:collaborativesynthesis} are then used to combine the fine-tuned models. Algorithm~\ref{alg:editing} displays the complete procedure of collaborative editing.
Note that fine-tuning the pre-trained models is for identity preservation proposed in Imagic, which is independent to our framework. The extension of our framework to editing requires no further training of \textit{dynamic diffusers}.

\algrenewcommand\algorithmicindent{0.5em}%
\begin{figure}[t]
\begin{minipage}[h]{0.492\textwidth}
\begin{algorithm}[H]
  \caption{Collaborative Editing} \label{alg:editing}
  \small
  \begin{algorithmic}[1]
    \vspace{.04in}
    \Require 
        \Statex input image $\bx_{input}$, target conditions $c_{m,target}$, 
        \Statex diffusion models $\bepsilon_{\theta_{m}}$, dynamic diffusers $\textbf{D}_{\phi_m}$, ($m = 1, \dotsc, M$), 
        \Statex interpolation scale $\alpha$
        \vspace{0.1cm}
    %
    %
    \For{$m = 1, ..., M$} \Comment{Uni-Modal Editing}
        \State $c_{m} = c_{m,target}$
        \State $\bx_{t} = \sqrt{\bar\alpha_t} \bx_{input} + \sqrt{1-\bar\alpha_t}\bepsilon$
        \State $c_{m,opt} = \operatorname*{\arg\!\min}_{c_{m}} \mathbb{E}_{\bepsilon, t} \left\| \bepsilon - \bepsilon_{\theta_{m}}(\bx_{t}, t, c_{m}) \right\|^2$
        \State $\theta_{m,opt} = \operatorname*{\arg\!\min}_{\theta_{m}} \mathbb{E}_{\bepsilon, t}  \left\| \bepsilon - \bepsilon_{\theta_{m}}(\bx_{t}, t, c_{m, opt}) \right\|^2$
        \State $c_{m,int} = \alpha \cdot c_{m,target} + (1 - \alpha) \cdot c_{m,opt}$
    \EndFor \vspace{0.1cm}
    \State $\bx_T \sim \mathcal{N}(\bzero, \bI)$ \Comment{Collaborate the Uni-Modal Edits}
    \For{$t=T, \dotsc, 1$} 
      \State $\bz \sim \mathcal{N}(\bzero, \bI)$ if $t > 1$, else $\bz = \bzero$  
      \For{$m = 0, ..., M$}
      \State \hskip1.0em $\bepsilon_{pred,m,t} = \bepsilon_{\theta_{m, opt}}(\bx_{t}, t, c_{m,int})$ \vspace{0.1cm}
      \State \hskip1.0em $\textbf{I}_{m,t} = \textbf{D}_{\phi_m}(\bx_{t}, t, c_{m,int})$ \vspace{0.1cm}
      \EndFor \vspace{0.1cm}
      \State $\hat{\textbf{I}}_{m,t, p} = \frac{\exp(\textbf{I}_{m,t, p})}{\sum_{j=1}^M \exp(\textbf{I}_{j,t, p})}$, softmax at each pixel $p$ \vspace{0.1cm} 
      \State $\bepsilon_{pred, t} = \sum_{m=1}^M \hat{\textbf{I}}_{m,t} \odot \bepsilon_{pred,m,t}$  \vspace{0.1cm}
      \State $\bx_{t-1} = \frac{1}{\sqrt{\alpha_t}}\left( \bx_t - \frac{1-\alpha_t}{\sqrt{1-\bar\alpha_t}} \bepsilon_{pred,t} \right) + \sigma_t \bz$
    \EndFor
    \State \textbf{return} $\bx_0$
    \vspace{.04in}
  \end{algorithmic}
\end{algorithm}
\end{minipage}
\vspace{-1em}
\end{figure}

\begin{figure}[t]
  \centering
    \includegraphics[width=0.99\linewidth]{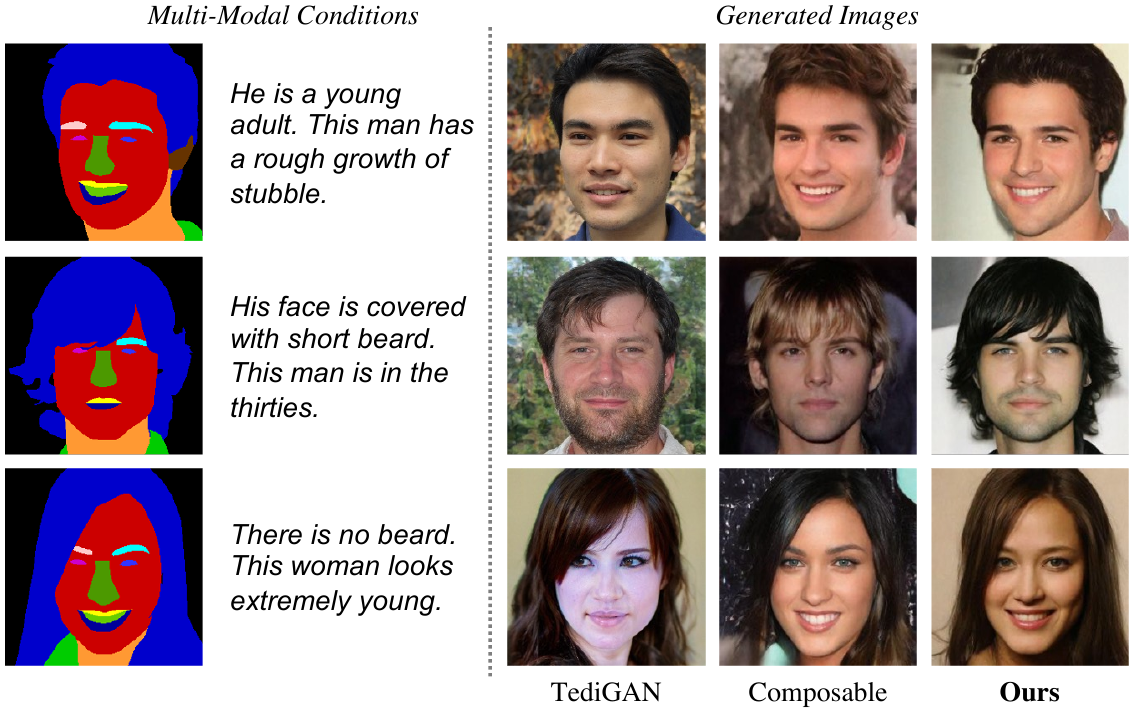}
    \vspace{-9pt}
   \caption{\textbf{Qualitative Comparison of Face Generation}. 
   In the second example, TediGAN and Composable fails to follow mask, while ours generates results highly consistent to both conditions.
   }
   \label{fig:baselinegeneration}
    \vspace{-5pt}
\end{figure}

\begin{figure}[t]
  \centering
    \includegraphics[width=0.99\linewidth]{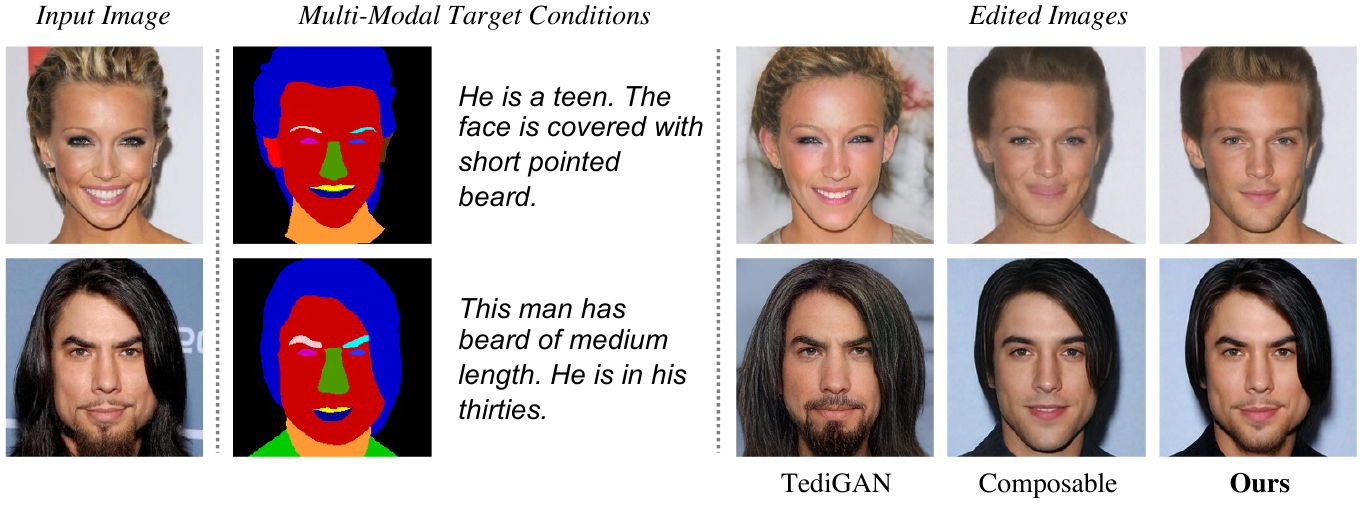}
    \vspace{-9pt}
   \caption{\textbf{Qualitative Comparison of Face Editing}. 
   While TediGAN's hair shape is inconsistent with mask, and Composable fails to generate beard according to text, our framework produces results highly consistent with both conditions while maintaining identity.
   }
   \label{fig:baselineediting}
    \vspace{-8pt}
\end{figure}

\section{Experiments}
\label{sec:experiments}

\subsection{Experimental Setup} 

\noindent\textbf{Datasets}. CelebAMask-HQ~\cite{lee2020maskgan} consists of manually annotated segmentation masks for each of the 30,000 images in the CelebA-HQ dataset~\cite{karras2018pggan}. Each mask has up to 19 classes, including the main facial components such as hair, skin, eyes, and nose, and accessories such as eyeglasses and cloth. CelebA-Dialog~\cite{jiang2021talktoedit} provides rich and fine-grained natural language descriptions for the images in CelebA-HQ. In this work, we train the mask-driven diffusion models on CelebAMask-HQ, and the text-driven models on CelebA-Dialog. For \textit{dynamic diffuser}, we simply combine the mask and text for each image without further processing.

\noindent\textbf{Implementation Details}.
We adopt LDM~\cite{rombach2022ldm} for our uni-modal diffusion models since it achieves a good balance between quality and speed. Our \textit{dynamic diffuser} is a time-conditional UNet~\cite{ronneberger2015unet}, with conditional embeddings injected via cross-attention~\cite{vaswani2017attention}. 
The \textit{dynamic diffuser} is $\times$30 smaller than the pre-trained diffusion model.
The detailed architecture and settings are provided in the supplementary material.

\begin{figure*}[t]
  \centering
    \includegraphics[width=0.99\textwidth, height=0.2\textwidth]{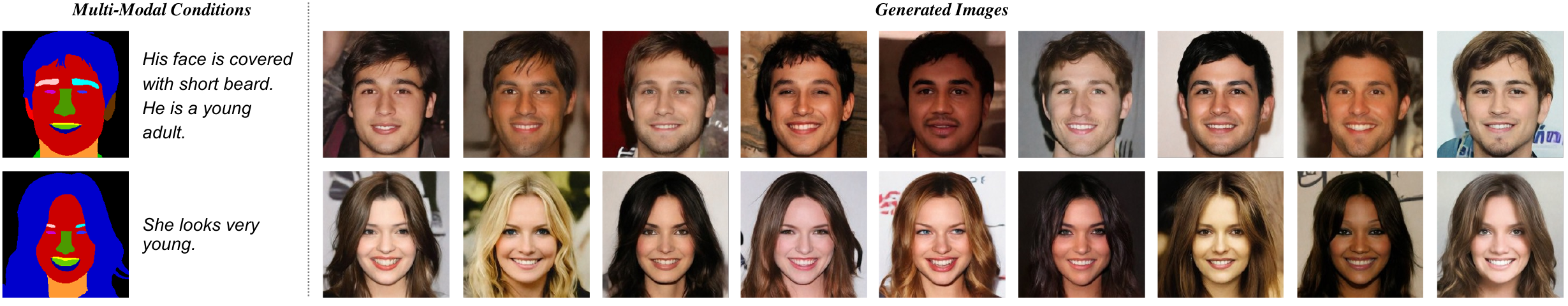}
   \caption{\textbf{Diversity of Face Generation}. Given input conditions, our method demonstrates promising synthesis diversity in facial attributes that are not constrained by a specific combination of multi-modal conditions, such as hair color and skin tone.}
   \label{fig:diversity}
\end{figure*}

\subsection{Comparison Methods} 
\noindent\textbf{TediGAN}~\cite{xia2021tedigan1, xia2021tedigan2} is a StyleGAN-based face synthesis and manipulation method. It projects both text and mask conditions into StyleGAN's $\mathcal{W}+$ latent space, and performs style mixing to achieve multi-modal control. 

\noindent\textbf{Composable Diffusion}~\cite{liu2022composablediffusion} parallely implements two instances of the same text-to-image diffusion model to perform compositional scene generation. We generalize it by replacing the two text-to-image model instances by the text-driven model and the mask-driven model trained using LDM~\cite{rombach2022ldm}. The two models are composed by averaging the intermediate diffusion results at each step of the \emph{reverse process}.

\subsection{Evaluation Metrics}

\noindent \textbf{FID}. 
We use Frechet Inception Distance (FID)~\cite{heusel2017fid} to measure the quality of images synthesized by different methods. FID computes the feature representation's distance between generated images and real images. A lower FID implies better sample quality.

\noindent \textbf{CLIP Score}. 
CLIP~\cite{radford2021clip} is a vision-language model trained on large-scale datasets. It uses an image encoder and a text encoder to project images and texts to a common feature space, respectively. The CLIP score is computed as the cosine similarity between the normalized image and text embeddings. A higher score usually indicates higher consistency between the output image and the text caption.

\noindent \textbf{Mask Accuracy}.
For each output image, we predict the segmentation mask using the face parsing network provided by CelebAMask-HQ~\cite{lee2020maskgan}. Mask accuracy is the pixel-wise accuracy against the ground-truth segmentation. A higher average accuracy indicates better consistency between the output image and the segmentation mask. 

\noindent \textbf{User Study}. 
We conduct user study with 25 human evaluators to measure the effectiveness of the methods perceptually. For \textit{face generation}, we randomly sample multi-modal conditions in the validation split of CelebA-HQ Dataset, and then synthesize output images given the conditions. 
Evaluators are provided with the input conditions and the output images, and they are asked to choose the best image based on 1) image photo-realism, 2) image-text consistency, 3) image-mask consistency. User study is conducted similarly for \textit{face editing}, except that 1) evaluators are also provided with the input image for editing, and 2) they are also asked to assess the level of identity preservation between the edited image and input image.

\begin{table}
  \centering
  
    \caption{\textbf{Quantitative Results of Face Generation}. 
    Compared with TediGAN and Composable Diffusion, our method synthesizes images with better quality (lower FID), and higher consistency with the text and mask conditions.
    }       
    \small
    \begin{tabular}{l|c|c|c}
    \Xhline{1pt}
    \textbf{Method} & \textbf{FID $\downarrow$} & \textbf{Text (\%) $\uparrow$} &\textbf{Mask (\%) $\uparrow$} \\ \Xhline{1pt}
    TediGAN~\cite{xia2021tedigan1, xia2021tedigan2} & 157.81 & 24.27 & 72.19 \\ 
    Composable~\cite{liu2022composablediffusion} & 124.62 & 23.94 & 76.11 \\ 
    \textbf{Ours} & \textbf{111.36} & \textbf{24.51} & \textbf{80.25} \\
    \Xhline{1pt}
  \end{tabular}
  
  \label{tab:example}
\end{table}

\begin{figure}[t]
  \centering
   \includegraphics[width=0.90\linewidth]{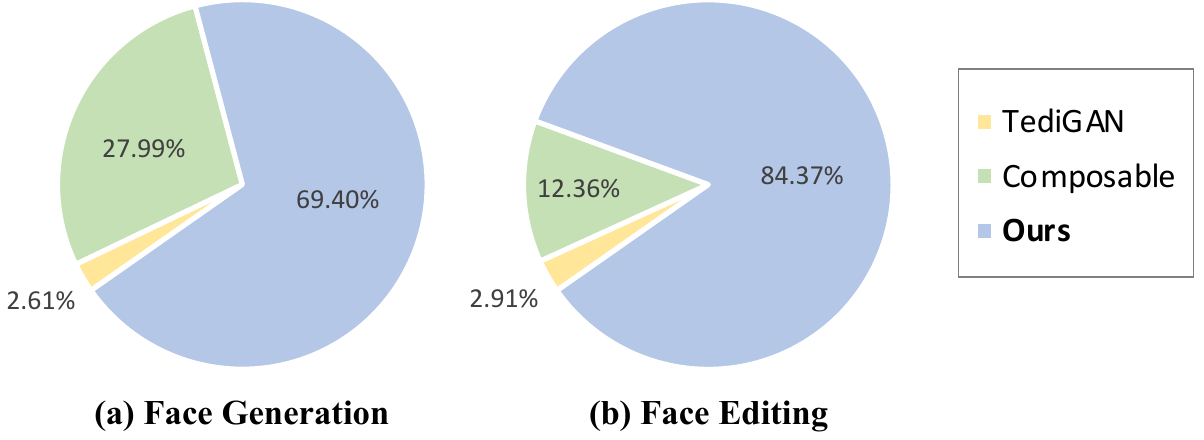}
   \caption{\textbf{User Study}. 
   Among the three methods, the majority of users vote for our results as the best for both generation (69.40\%) and editing (84.37\%), in terms of image quality, condition consistency, and identity preservation.
   }
   \label{fig:userstudy}
\end{figure}

\subsection{Comparative Study}
\begin{figure*}[t]
  \centering
    \includegraphics[width=0.99\textwidth]{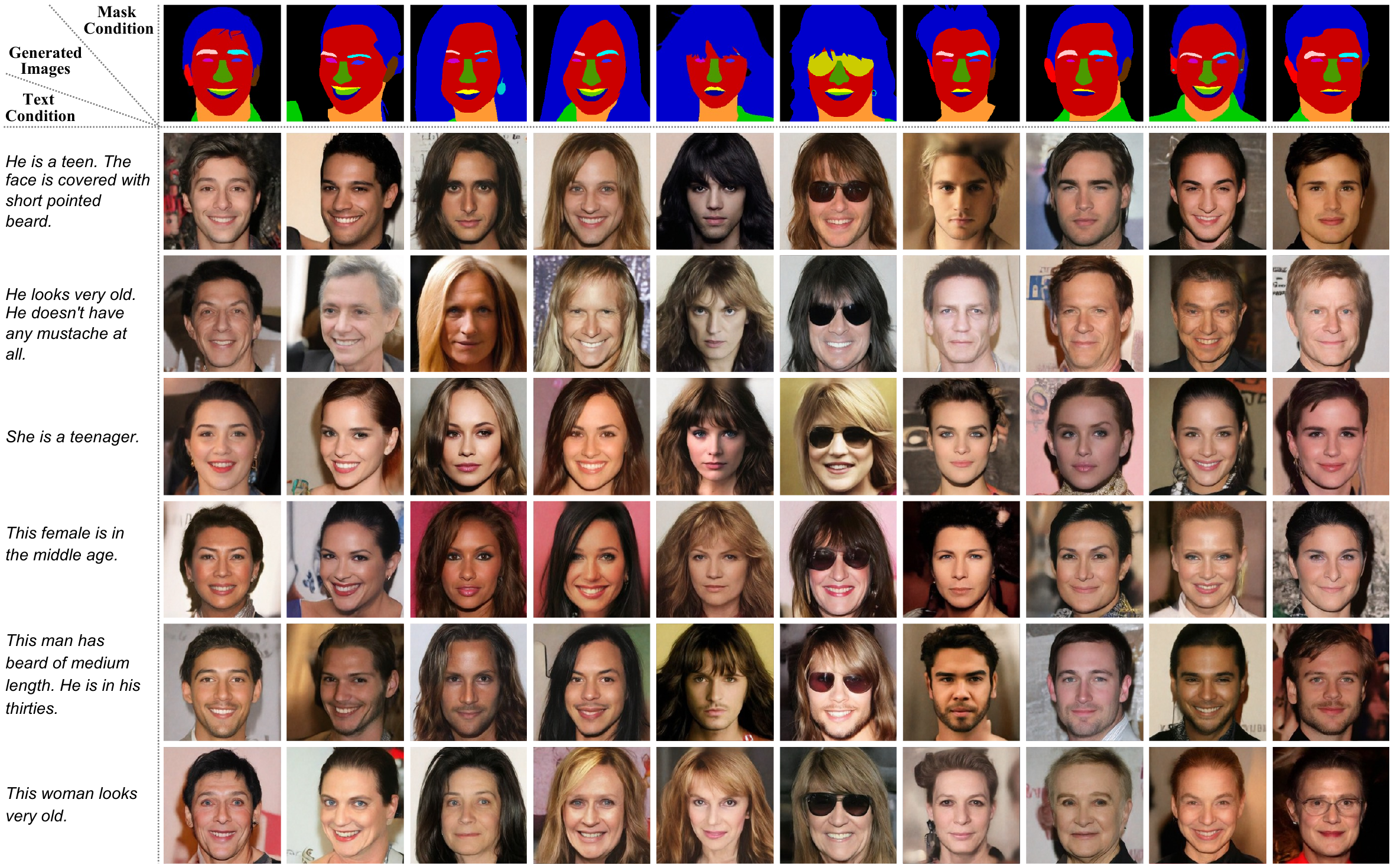}
   \caption{\textbf{Face Generation Results}. Our method generates realistic images under different combinations of multi-modal conditions, even for relatively rare combinations in the training distribution, such as a man with long hair.}
   \label{fig:generation}
\end{figure*}

\noindent\textbf{Qualitative Comparison.}
In Figure~\ref{fig:generation}, we provide the generation results of our \textit{Collaborative Diffusion} under different combinations of texts and masks. It is observed that our framework is capable of synthesizing realistic outputs consistent with the multi-modal inputs, even for relatively rare conditions, such as a man with long hair.
We then compare our method with TediGAN~\cite{xia2021tedigan1, xia2021tedigan2} and Composable Diffusion~\cite{liu2022composablediffusion} in Figure~\ref{fig:baselinegeneration}. For example, in the second example, TediGAN and Composable Diffusion's results are inconsistent with the mask, while ours is able to maintain both mask and text consistency. We also show in Figure~\ref{fig:diversity} that our method synthesizes images with high diversity without losing condition consistency.

\begin{figure*}[t]
  \centering
    \includegraphics[width=0.95\textwidth]{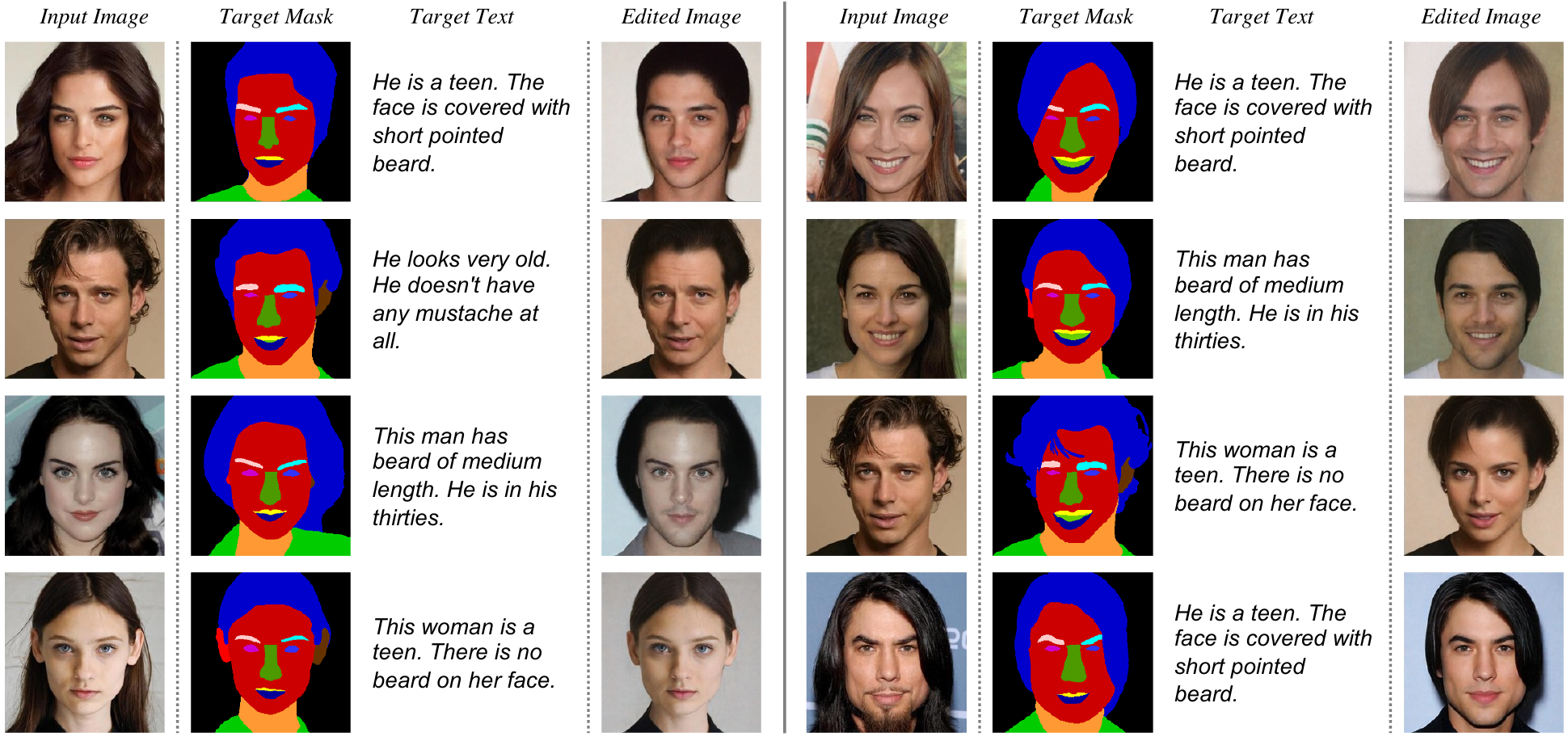}
   \caption{\textbf{Face Editing Results}. Given the input real image and target conditions, we display the edited image using our method.}
   \label{fig:editing}
\end{figure*}

In addition to image generation, it is also observed in Figure~\ref{fig:editing} that our framework is able to edit images based on multi-modal conditions while preserving identity. The comparison to existing works in Figure~\ref{fig:baselineediting} also verifies our effectiveness. For example, in the second example, while TediGAN is unable to synthesize hair consistent to the mask and Composable Diffusion fails to generate beard according to the text, our framework is able to generate results highly consistent to both conditions while maintaining identity.

\noindent\textbf{Quantitative Comparison.}
As shown in Table~\ref{tab:example}, our \textit{Collaborate Diffusion} outperforms TediGAN and Composable Diffusion in all three objective metrics. In addition, as depicted in Figure~\ref{fig:userstudy}, our framework achieves the best result in 69.40\% and 84.37\% of the time for face generation and editing, respectively. These results verify our effectiveness in both image quality and image-condition consistency.

\subsection{Ablation Study}
\label{sec:ablationstudy}
In this section, we visualize the \emph{influence functions}, and show that it is necessary for the \emph{influence function} to be both spatially and temporally varying to facilitate effective collaboration.

\subsubsection{Spatial Variations of Influence Functions}
\begin{figure}[t]
  \centering
  \includegraphics[width=0.99\linewidth]{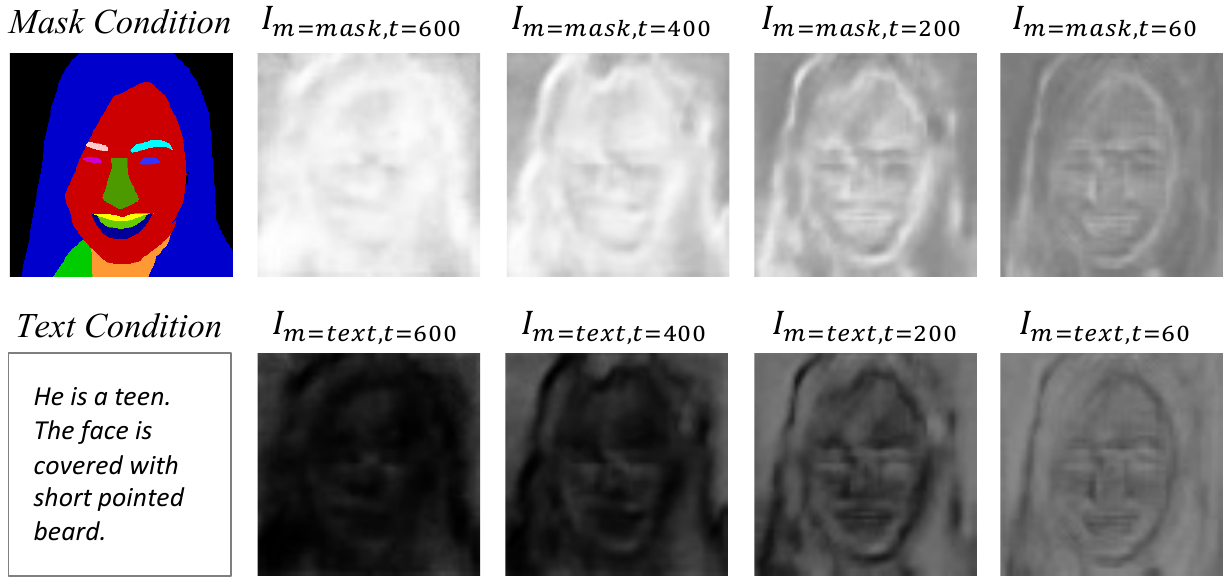}
  \caption{\textbf{Visualization of Influence Functions}. 
  The \textit{influence functions} vary spatially at different face regions, and temporally at different diffusion timesteps.
  }
  \label{fig:influencefunctions}
  \vspace{-5pt}
\end{figure}

As shown in Figure~\ref{fig:influencefunctions}, the \textit{influence functions} behaves differently at different spatial regions. For instance, the influence for the mask-driven model mainly lies on the contours of facial regions, such as the outline of hair, face, and eyes, as these regions are crucial in defining facial layout. In contrast, the influence for the text-driven model is stronger at skin regions including cheeks and chin. This is because the attributes related skin texture, such as age and beard length, are better described by text.

To verify its necessity, we remove the spatial variation of \emph{influence functions}. From Table~\ref{tab:ablation} we observe that removing the spatial variance results in deterioration in both output quality and condition consistency. This corroborates our hypothesis that it is important to assign different weights to different modalities at different spatial regions.

\subsubsection{Temporal Variations of Influence Functions}
In addition to the spatial variation, it is also observed in Figure~\ref{fig:influencefunctions} that the influence from the mask-driven model is stronger at earlier diffusion stages (\ie, larger $t$), since early stages focus on initializing the facial layout using the mask-driven model's predictions. At later stages, the influence from the text-driven model increases as the textural details (\eg, skin wrinkles and beard length) are instantiated using information from the text. 

We remove the temporal variation in the \textit{influence functions} to demonstrate its importance. It is shown in Table~\ref{tab:ablation} that both the quality and condition consistency drop without the temporal variation. 

\begin{table}[t]
  \centering
  \caption{\textbf{Ablation Study}. 
  Temporal or spatial suppression in influence variation introduces performance drops, which shows the \textit{necessity of influence functions' spatial-temporal adaptivity}.
  }
  \vspace{-0.5em}
    \small
    \begin{tabular}{l|c|c|c}
    \Xhline{1pt}
    \textbf{Method} & \textbf{FID $\downarrow$} & \textbf{Text (\%) $\uparrow$} & \textbf{Mask (\%) $\uparrow$} \\ \Xhline{1pt}
    Ours w/o Spatial & 117.81 & 24.36 & 80.08 \\ 
    Ours w/o Temporal & 117.34 & 24.48 & 77.07 \\ 
    \textbf{Ours} & \textbf{111.36} & \textbf{24.51} & \textbf{80.25} \\
    \Xhline{1pt}
  \end{tabular}
  \vspace{-5pt}
  \label{tab:ablation}
\end{table}

\section{Conclusion}

With generative AI gaining increasing attention, multi-modal conditioning becomes an indispensable direction to unleash creativity and enable comprehensive controls from human creators. In this work, we take the first step forward and propose \textbf{\textit{Collaborative Diffusion}}, where pre-trained uni-modal diffusion models collaboratively achieve multi-modal face generation and editing without being re-trained. With our \textit{dynamic diffuser}, this framework could be used to extend arbitrary uni-modal approach to the multi-modal paradigm through predicting the relative influence of different modalities. We believe our idea of synergizing uni-modal models for multi-modal tasks would be a good inspiration for future works in various domains such as motion and 3D generation.

\noindent \textbf{Limitations and Future Work}.
Since our framework focuses on exploiting pre-trained diffusion models, our performance is dependent on the capability of each model. An orthogonal direction is to train each collaborator on large-scale datasets for performance gain. 

\noindent\textbf{Potential Negative Societal Impacts}.
The facial manipulation capabilities of \textit{Collaborative Diffusion} could be applied maliciously on real human faces. We advise users to apply it only for proper recreational purposes.

\noindent\textbf{Acknowledgement}.
We would like to thank Liangyu Chen and Bo Li for proofreading.
This study is supported by NTU NAP, MOE AcRF Tier 1 (2021-T1-001-088), and under the RIE2020 Industry Alignment Fund – Industry Collaboration Projects (IAF-ICP) Funding Initiative, as well as cash and in-kind contribution from the industry partner(s).

{\small
\bibliographystyle{ieee_fullname}
\bibliography{main}

\begin{thebibliography}{10}\itemsep=-1pt

\bibitem{amit2021segdiff}
Tomer Amit, Eliya Nachmani, Tal Shaharbany, and Lior Wolf.
\newblock {SegDiff}: Image segmentation with diffusion probabilistic models.
\newblock {\em arXiv preprint arXiv:2112.00390}, 2021.

\bibitem{austin2021structured}
Jacob Austin, Daniel~D Johnson, Jonathan Ho, Daniel Tarlow, and Rianne van~den
  Berg.
\newblock Structured denoising diffusion models in discrete state-spaces.
\newblock In {\em NeurIPS}, 2021.

\bibitem{avrahami2022blended}
Omri Avrahami, Dani Lischinski, and Ohad Fried.
\newblock Blended diffusion for text-driven editing of natural images.
\newblock In {\em CVPR}, 2022.

\bibitem{ba2016layer}
Jimmy~Lei Ba, Jamie~Ryan Kiros, and Geoffrey~E Hinton.
\newblock Layer normalization.
\newblock {\em arXiv preprint arXiv:1607.06450}, 2016.

\bibitem{baranchuk2021label}
Dmitry Baranchuk, Ivan Rubachev, Andrey Voynov, Valentin Khrulkov, and Artem
  Babenko.
\newblock Label-efficient semantic segmentation with diffusion models.
\newblock In {\em ICLR}, 2022.

\bibitem{bond2022unleashing}
Sam Bond-Taylor, Peter Hessey, Hiroshi Sasaki, Toby~P Breckon, and Chris~G
  Willcocks.
\newblock Unleashing transformers: parallel token prediction with discrete
  absorbing diffusion for fast high-resolution image generation from
  vector-quantized codes.
\newblock In {\em ECCV}, 2022.

\bibitem{chang2022maskgit}
Huiwen Chang, Han Zhang, Lu Jiang, Ce Liu, and William~T Freeman.
\newblock {MaskGIT}: Masked generative image transformer.
\newblock In {\em CVPR}, 2022.

\bibitem{cheng2023adaptively}
Shin-I Cheng, Yu-Jie Chen, Wei-Chen Chiu, Hung-Yu Tseng, and Hsin-Ying Lee.
\newblock Adaptively-realistic image generation from stroke and sketch with
  diffusion model.
\newblock In {\em WACV}, 2023.

\bibitem{couairon2022diffedit}
Guillaume Couairon, Jakob Verbeek, Holger Schwenk, and Matthieu Cord.
\newblock Diffedit: Diffusion-based semantic image editing with mask guidance.
\newblock {\em arXiv preprint arXiv:2210.11427}, 2022.

\bibitem{devlin2018bert}
Jacob Devlin, Ming-Wei Chang, Kenton Lee, and Kristina Toutanova.
\newblock {BERT}: Pre-training of deep bidirectional transformers for language
  understanding.
\newblock {\em arXiv preprint arXiv:1810.04805}, 2018.

\bibitem{dhariwal2021beatgan}
Prafulla Dhariwal and Alexander Nichol.
\newblock Diffusion models beat {GAN}s on image synthesis.
\newblock In {\em NeurIPS}, 2021.

\bibitem{ding2021cogview}
Ming Ding, Zhuoyi Yang, Wenyi Hong, Wendi Zheng, Chang Zhou, Da Yin, Junyang
  Lin, Xu Zou, Zhou Shao, Hongxia Yang, et~al.
\newblock {CogView}: Mastering text-to-image generation via transformers.
\newblock In {\em NeurIPS}, 2021.

\bibitem{esser2021imagebart}
Patrick Esser, Robin Rombach, Andreas Blattmann, and Bjorn Ommer.
\newblock {ImageBART}: Bidirectional context with multinomial diffusion for
  autoregressive image synthesis.
\newblock In {\em NeurIPS}, 2021.

\bibitem{esser2020note}
Patrick Esser, Robin Rombach, and Bj{\"o}rn Ommer.
\newblock A note on data biases in generative models.
\newblock In {\em NeurIPS Workshop}, 2020.

\bibitem{esser2021vqgan}
Patrick Esser, Robin Rombach, and Bjorn Ommer.
\newblock Taming transformers for high-resolution image synthesis.
\newblock In {\em CVPR}, 2021.

\bibitem{gafni2022makeascene}
Oran Gafni, Adam Polyak, Oron Ashual, Shelly Sheynin, Devi Parikh, and Yaniv
  Taigman.
\newblock Make-a-scene: Scene-based text-to-image generation with human priors.
\newblock {\em arXiv preprint arXiv:2203.13131}, 2022.

\bibitem{goodfellow2014gan}
Ian~J Goodfellow, Jean Pouget-Abadie, Mehdi Mirza, Bing Xu, David Warde-Farley,
  Sherjil Ozair, Aaron~C Courville, and Yoshua Bengio.
\newblock Generative adversarial nets.
\newblock In {\em NeurIPS}, 2014.

\bibitem{graikos2022diffusion}
Alexandros Graikos, Nikolay Malkin, Nebojsa Jojic, and Dimitris Samaras.
\newblock Diffusion models as plug-and-play priors.
\newblock In {\em NeurIPS}, 2022.

\bibitem{gu2022vqdiffusion}
Shuyang Gu, Dong Chen, Jianmin Bao, Fang Wen, Bo Zhang, Dongdong Chen, Lu Yuan,
  and Baining Guo.
\newblock Vector quantized diffusion model for text-to-image synthesis.
\newblock In {\em CVPR}, 2022.

\bibitem{harvey2022fdm}
William Harvey, Saeid Naderiparizi, Vaden Masrani, Christian Weilbach, and
  Frank Wood.
\newblock Flexible diffusion modeling of long videos.
\newblock {\em arXiv preprint arXiv:2205.11495}, 2022.

\bibitem{heusel2017fid}
Martin Heusel, Hubert Ramsauer, Thomas Unterthiner, Bernhard Nessler, and Sepp
  Hochreiter.
\newblock {GANs} trained by a two time-scale update rule converge to a local
  nash equilibrium.
\newblock In {\em NeurIPS}, 2017.

\bibitem{ho2022imagenvideo}
Jonathan Ho, William Chan, Chitwan Saharia, Jay Whang, Ruiqi Gao, Alexey
  Gritsenko, Diederik~P Kingma, Ben Poole, Mohammad Norouzi, David~J Fleet,
  et~al.
\newblock Imagen video: High definition video generation with diffusion models.
\newblock {\em arXiv preprint arXiv:2210.02303}, 2022.

\bibitem{ho2020ddpm}
Jonathan Ho, Ajay Jain, and Pieter Abbeel.
\newblock Denoising diffusion probabilistic models.
\newblock In {\em NeurIPS}, 2020.

\bibitem{ho2022cascaded}
Jonathan Ho, Chitwan Saharia, William Chan, David~J Fleet, Mohammad Norouzi,
  and Tim Salimans.
\newblock Cascaded diffusion models for high fidelity image generation.
\newblock {\em JMLR}, 2022.

\bibitem{hyvarinen2005estimation}
Aapo Hyv{\"a}rinen and Peter Dayan.
\newblock Estimation of non-normalized statistical models by score matching.
\newblock {\em JMLR}, 2005.

\bibitem{jiang2021talktoedit}
Yuming Jiang, Ziqi Huang, Xingang Pan, Chen~Change Loy, and Ziwei Liu.
\newblock {Talk-to-Edit}: Fine-grained facial editing via dialog.
\newblock In {\em ICCV}, 2021.

\bibitem{jiang2022text2human}
Yuming Jiang, Shuai Yang, Haonan Qju, Wayne Wu, Chen~Change Loy, and Ziwei Liu.
\newblock Text2human: Text-driven controllable human image generation.
\newblock {\em ACM TOG}, 2022.

\bibitem{johnson2016perceptual}
Justin Johnson, Alexandre Alahi, and Fei-Fei Li.
\newblock Perceptual losses for real-time style transfer and super-resolution.
\newblock In {\em ECCV}, 2016.

\bibitem{karras2018pggan}
Tero Karras, Timo Aila, Samuli Laine, and Jaakko Lehtinen.
\newblock Progressive growing of {GAN}s for improved quality, stability, and
  variation.
\newblock In {\em ICLR}, 2018.

\bibitem{Karras2021stylegan3}
Tero Karras, Miika Aittala, Samuli Laine, Erik H{\"a}rk{\"o}nen, Janne
  Hellsten, Jaakko Lehtinen, and Timo Aila.
\newblock Alias-free generative adversarial networks.
\newblock In {\em NeurIPS}, 2021.

\bibitem{Karras2019stylegan1}
Tero Karras, Samuli Laine, and Timo Aila.
\newblock A style-based generator architecture for generative adversarial
  networks.
\newblock In {\em CVPR}, 2019.

\bibitem{karras2020stylegan2}
Tero Karras, Samuli Laine, Miika Aittala, Janne Hellsten, Jaakko Lehtinen, and
  Timo Aila.
\newblock Analyzing and improving the image quality of {StyleGAN}.
\newblock In {\em CVPR}, 2020.

\bibitem{kawar2022imagic}
Bahjat Kawar, Shiran Zada, Oran Lang, Omer Tov, Huiwen Chang, Tali Dekel, Inbar
  Mosseri, and Michal Irani.
\newblock Imagic: Text-based real image editing with diffusion models.
\newblock {\em arXiv preprint arXiv:2210.09276}, 2022.

\bibitem{kim2022diffusionclip}
Gwanghyun Kim, Taesung Kwon, and Jong~Chul Ye.
\newblock {DiffusionCLIP}: Text-guided diffusion models for robust image
  manipulation.
\newblock In {\em CVPR}, 2022.

\bibitem{kingma2013vae}
Diederik~P Kingma and Max Welling.
\newblock Auto-encoding variational bayes.
\newblock {\em arXiv preprint arXiv:1312.6114}, 2013.

\bibitem{koh2021text}
Jing~Yu Koh, Jason Baldridge, Honglak Lee, and Yinfei Yang.
\newblock Text-to-image generation grounded by fine-grained user attention.
\newblock In {\em WACV}, 2021.

\bibitem{lee2020maskgan}
Cheng-Han Lee, Ziwei Liu, Lingyun Wu, and Ping Luo.
\newblock {MaskGAN}: Towards diverse and interactive facial image manipulation.
\newblock In {\em CVPR}, 2020.

\bibitem{li2020manigan}
Bowen Li, Xiaojuan Qi, Thomas Lukasiewicz, and Philip~HS Torr.
\newblock {ManiGAN}: Text-guided image manipulation.
\newblock In {\em CVPR}, 2020.

\bibitem{li2019object}
Wenbo Li, Pengchuan Zhang, Lei Zhang, Qiuyuan Huang, Xiaodong He, Siwei Lyu,
  and Jianfeng Gao.
\newblock Object-driven text-to-image synthesis via adversarial training.
\newblock In {\em CVPR}, 2019.

\bibitem{li2022stylet2i}
Zhiheng Li, Martin~Renqiang Min, Kai Li, and Chenliang Xu.
\newblock {StyleT2I}: Toward compositional and high-fidelity text-to-image
  synthesis.
\newblock In {\em CVPR}, 2022.

\bibitem{liu2022composablediffusion}
Nan Liu, Shuang Li, Yilun Du, Antonio Torralba, and Joshua~B Tenenbaum.
\newblock Compositional visual generation with composable diffusion models.
\newblock {\em arXiv preprint arXiv:2206.01714}, 2022.

\bibitem{liu2023more}
Xihui Liu, Dong~Huk Park, Samaneh Azadi, Gong Zhang, Arman Chopikyan, Yuxiao
  Hu, Humphrey Shi, Anna Rohrbach, and Trevor Darrell.
\newblock More control for free! image synthesis with semantic diffusion
  guidance.
\newblock In {\em WACV}, 2023.

\bibitem{meng2021sdedit}
Chenlin Meng, Yutong He, Yang Song, Jiaming Song, Jiajun Wu, Jun-Yan Zhu, and
  Stefano Ermon.
\newblock {SDEdit}: Guided image synthesis and editing with stochastic
  differential equations.
\newblock In {\em ICLR}, 2022.

\bibitem{nichol2021glide}
Alex Nichol, Prafulla Dhariwal, Aditya Ramesh, Pranav Shyam, Pamela Mishkin,
  Bob McGrew, Ilya Sutskever, and Mark Chen.
\newblock {GLIDE}: Towards photorealistic image generation and editing with
  text-guided diffusion models.
\newblock {\em arXiv preprint arXiv:2112.10741}, 2021.

\bibitem{park2019spade}
Taesung Park, Ming-Yu Liu, Ting-Chun Wang, and Jun-Yan Zhu.
\newblock Semantic image synthesis with spatially-adaptive normalization.
\newblock In {\em CVPR}, 2019.

\bibitem{patashnik2021styleclip}
Or Patashnik, Zongze Wu, Eli Shechtman, Daniel Cohen-Or, and Dani Lischinski.
\newblock {StyleCLIP}: Text-driven manipulation of stylegan imagery.
\newblock In {\em ICCV}, 2021.

\bibitem{radford2021clip}
Alec Radford, Jong~Wook Kim, Chris Hallacy, Aditya Ramesh, Gabriel Goh,
  Sandhini Agarwal, Girish Sastry, Amanda Askell, Pamela Mishkin, Jack Clark,
  et~al.
\newblock Learning transferable visual models from natural language
  supervision.
\newblock In {\em ICML}, 2021.

\bibitem{ramesh2022dalle2}
Aditya Ramesh, Prafulla Dhariwal, Alex Nichol, Casey Chu, and Mark Chen.
\newblock Hierarchical text-conditional image generation with {CLIP} latents.
\newblock {\em arXiv preprint arXiv:2204.06125}, 2022.

\bibitem{ramesh2021dalle1}
Aditya Ramesh, Mikhail Pavlov, Gabriel Goh, Scott Gray, Chelsea Voss, Alec
  Radford, Mark Chen, and Ilya Sutskever.
\newblock Zero-shot text-to-image generation.
\newblock In {\em ICML}, 2021.

\bibitem{reed2016generative}
Scott Reed, Zeynep Akata, Xinchen Yan, Lajanugen Logeswaran, Bernt Schiele, and
  Honglak Lee.
\newblock Generative adversarial text to image synthesis.
\newblock In {\em ICML}, 2016.

\bibitem{reed2016learning}
Scott~E Reed, Zeynep Akata, Santosh Mohan, Samuel Tenka, Bernt Schiele, and
  Honglak Lee.
\newblock Learning what and where to draw.
\newblock In {\em NeurIPS}, 2016.

\bibitem{richardson2021encoding}
Elad Richardson, Yuval Alaluf, Or Patashnik, Yotam Nitzan, Yaniv Azar, Stav
  Shapiro, and Daniel Cohen-Or.
\newblock Encoding in style: a {StyleGAN} encoder for image-to-image
  translation.
\newblock In {\em CVPR}, 2021.

\bibitem{rombach2022ldm}
Robin Rombach, Andreas Blattmann, Dominik Lorenz, Patrick Esser, and Bj{\"o}rn
  Ommer.
\newblock High-resolution image synthesis with latent diffusion models.
\newblock In {\em CVPR}, 2022.

\bibitem{ronneberger2015unet}
Olaf Ronneberger, Philipp Fischer, and Thomas Brox.
\newblock {U-Net}: Convolutional networks for biomedical image segmentation.
\newblock In {\em MICCAI}, 2015.

\bibitem{ruiz2022dreambooth}
Nataniel Ruiz, Yuanzhen Li, Varun Jampani, Yael Pritch, Michael Rubinstein, and
  Kfir Aberman.
\newblock Dreambooth: Fine tuning text-to-image diffusion models for
  subject-driven generation.
\newblock {\em arXiv preprint arXiv:2208.12242}, 2022.

\bibitem{saharia2022imagen}
Chitwan Saharia, William Chan, Saurabh Saxena, Lala Li, Jay Whang, Emily
  Denton, Seyed Kamyar~Seyed Ghasemipour, Burcu~Karagol Ayan, S~Sara Mahdavi,
  Rapha~Gontijo Lopes, et~al.
\newblock Photorealistic text-to-image diffusion models with deep language
  understanding.
\newblock {\em arXiv preprint arXiv:2205.11487}, 2022.

\bibitem{saharia2022sr3}
Chitwan Saharia, Jonathan Ho, William Chan, Tim Salimans, David~J Fleet, and
  Mohammad Norouzi.
\newblock Image super-resolution via iterative refinement.
\newblock {\em IEEE TPAMI}, 2022.

\bibitem{shen2020interfacegan1}
Yujun Shen, Jinjin Gu, Xiaoou Tang, and Bolei Zhou.
\newblock Interpreting the latent space of {GAN}s for semantic face editing.
\newblock In {\em CVPR}, 2020.

\bibitem{shen2020interfacegan2}
Yujun Shen, Ceyuan Yang, Xiaoou Tang, and Bolei Zhou.
\newblock {InterFaceGAN}: Interpreting the disentangled face representation
  learned by {GAN}s.
\newblock {\em IEEE TPAMI}, 2020.

\bibitem{simonyan2014vgg}
Karen Simonyan and Andrew Zisserman.
\newblock Very deep convolutional networks for large-scale image recognition.
\newblock {\em arXiv preprint arXiv:1409.1556}, 2014.

\bibitem{singer2022makeavideo}
Uriel Singer, Adam Polyak, Thomas Hayes, Xi Yin, Jie An, Songyang Zhang, Qiyuan
  Hu, Harry Yang, Oron Ashual, Oran Gafni, et~al.
\newblock Make-a-video: Text-to-video generation without text-video data.
\newblock {\em arXiv preprint arXiv:2209.14792}, 2022.

\bibitem{sohl2015deep}
Jascha Sohl-Dickstein, Eric Weiss, Niru Maheswaranathan, and Surya Ganguli.
\newblock Deep unsupervised learning using nonequilibrium thermodynamics.
\newblock In {\em ICML}, 2015.

\bibitem{song2020ddim}
Jiaming Song, Chenlin Meng, and Stefano Ermon.
\newblock Denoising diffusion implicit models.
\newblock In {\em ICLR}, 2021.

\bibitem{song2020score}
Yang Song, Jascha Sohl-Dickstein, Diederik~P Kingma, Abhishek Kumar, Stefano
  Ermon, and Ben Poole.
\newblock Score-based generative modeling through stochastic differential
  equations.
\newblock In {\em ICLR}, 2021.

\bibitem{song2018talking}
Yang Song, Jingwen Zhu, Dawei Li, Xiaolong Wang, and Hairong Qi.
\newblock Talking face generation by conditional recurrent adversarial network.
\newblock {\em IJCAI}, 2019.

\bibitem{sun2022anyface}
Jianxin Sun, Qiyao Deng, Qi Li, Muyi Sun, Min Ren, and Zhenan Sun.
\newblock {AnyFace}: Free-style text-to-face synthesis and manipulation.
\newblock In {\em CVPR}, 2022.

\bibitem{tinsley2021face}
Patrick Tinsley, Adam Czajka, and Patrick Flynn.
\newblock This face does not exist... but it might be yours! identity leakage
  in generative models.
\newblock In {\em WACV}, 2021.

\bibitem{van2017vqvae}
Aaron Van Den~Oord, Oriol Vinyals, et~al.
\newblock Neural discrete representation learning.
\newblock In {\em NeurIPS}, 2017.

\bibitem{vaswani2017attention}
Ashish Vaswani, Noam Shazeer, Niki Parmar, Jakob Uszkoreit, Llion Jones,
  Aidan~N Gomez, {\L}ukasz Kaiser, and Illia Polosukhin.
\newblock Attention is all you need.
\newblock In {\em NeurIPS}, 2017.

\bibitem{villegas2022phenaki}
Ruben Villegas, Mohammad Babaeizadeh, Pieter-Jan Kindermans, Hernan Moraldo,
  Han Zhang, Mohammad~Taghi Saffar, Santiago Castro, Julius Kunze, and Dumitru
  Erhan.
\newblock Phenaki: Variable length video generation from open domain textual
  description.
\newblock {\em arXiv preprint arXiv:2210.02399}, 2022.

\bibitem{vincent2011connection}
Pascal Vincent.
\newblock A connection between score matching and denoising autoencoders.
\newblock {\em Neural Computation}, 2011.

\bibitem{wang2022piti}
Tengfei Wang, Ting Zhang, Bo Zhang, Hao Ouyang, Dong Chen, Qifeng Chen, and
  Fang Wen.
\newblock Pretraining is all you need for image-to-image translation.
\newblock {\em arXiv preprint arXiv:2205.12952}, 2022.

\bibitem{wang2022semantic}
Weilun Wang, Jianmin Bao, Wengang Zhou, Dongdong Chen, Dong Chen, Lu Yuan, and
  Houqiang Li.
\newblock Semantic image synthesis via diffusion models.
\newblock {\em arXiv preprint arXiv:2207.00050}, 2022.

\bibitem{xia2021tedigan1}
Weihao Xia, Yujiu Yang, Jing-Hao Xue, and Baoyuan Wu.
\newblock {TediGAN}: Text-guided diverse face image generation and
  manipulation.
\newblock In {\em CVPR}, 2021.

\bibitem{xia2021tedigan2}
Weihao Xia, Yujiu Yang, Jing-Hao Xue, and Baoyuan Wu.
\newblock Towards open-world text-guided face image generation and
  manipulation.
\newblock {\em arXiv preprint arXiv:2104.08910}, 2021.

\bibitem{xu2018attngan}
Tao Xu, Pengchuan Zhang, Qiuyuan Huang, Han Zhang, Zhe Gan, Xiaolei Huang, and
  Xiaodong He.
\newblock {AttnGAN}: Fine-grained text to image generation with attentional
  generative adversarial networks.
\newblock In {\em CVPR}, 2018.

\bibitem{zhang2017stackgan}
Han Zhang, Tao Xu, Hongsheng Li, Shaoting Zhang, Xiaogang Wang, Xiaolei Huang,
  and Dimitris~N Metaxas.
\newblock {StackGAN}: Text to photo-realistic image synthesis with stacked
  generative adversarial networks.
\newblock In {\em ICCV}, 2017.

\bibitem{zhang2018stackgan++}
Han Zhang, Tao Xu, Hongsheng Li, Shaoting Zhang, Xiaogang Wang, Xiaolei Huang,
  and Dimitris~N Metaxas.
\newblock {StackGAN++}: Realistic image synthesis with stacked generative
  adversarial networks.
\newblock {\em IEEE TPAMI}, 2018.

\end{thebibliography}
}


\onecolumn
\appendix
\section*{Supplementary}
\renewcommand\thesection{\Alph{section}}
\renewcommand\thefigure{A\arabic{figure}}
\renewcommand\thetable{A\arabic{table}}
\setlength{\parskip}{3pt}

In this \textbf{\textit{supplementary file}}, we elaborate on the implementation details of the \textit{Collaborative Diffusion} framework in Section~\ref{sec:implementation_details}. We then provide further explanations on experimental details in  Section~\ref{sec:experimental}. More qualitative results and visualizations are provided in Section~\ref{sec:qualitative}. Finally, we discuss the potential societal impacts in Section~\ref{sec:impact}.
%
%
\section{Implementation Details}
\label{sec:implementation_details}

In this section, we describe the implementation details of our \textbf{\textit{Collaborative Diffusion}} framework.

\subsection{Multi-Modal Collaborative Synthesis}

We adopt LDM~\cite{rombach2022ldm} as our uni-modal diffusion models for its good balance between quality and speed. LDM~\cite{rombach2022ldm} applies diffusion models in the latent space of autoencoders to reduce the computation overhead of training and sampling. Our framework supports both $256{\times}256$ and $512{\times}512$ resolution, and we will use the $256{\times}256$ version in subsequent discussions for simplicity.

We train a Variational Autoencoder (VAE)~\cite{kingma2013vae}, where the encoder compresses $256{\times}256{\times}3$ resolution images into the $64{\times}64{\times}3$ latent space, and the decoder reconstructs the $256{\times}256{\times}3$ images from the $64{\times}64{\times}3$ latent codes. The VAE is trained on the CelebA-HQ~\cite{karras2018pggan} Dataset by minimizing the following objective:
\begin{gather}
L_{\textit{VAE}} = 1.0 \cdot L_{rec} + 1.0 \cdot L_{vgg} + 10^{-6} \cdot L_{kl},
\end{gather}
where $L_{rec}$ is the $L_1$ distance between the reconstructed image and the input image, $L_{vgg}$ is the perceptual loss~\cite{johnson2016perceptual} using VGG-16~\cite{simonyan2014vgg}, and $L_{kl}$ is the Kullback–Leibler divergence term which regularizes the VAE latent space towards the Gaussian distribution. The KL term is largely scaled down by a factor of $10^{-6}$ for two reasons: 1) KL regularization was required in the original VAE for directly sampling latent codes from the Gaussian prior. In this work, we simply use VAE as an image compression tool instead of a generative model, so that we do not need strong regularization of VAE latent space. Diffusion models will take care of sampling meaningful latent codes from the weakly regularized VAE latent space. 2) Weaker KL regularization allows relatively stronger focus on image reconstruction, and thus potentially less distortion during VAE's compression-reconstruction process. 
All our  \textit{dynamic diffusers} and uni-modal diffusion models are applied in the $64{\times}64{\times}3$ latent space of the pre-trained VAE. The \textit{reverse process} of diffusion models gradually denoises the Gaussian $\textbf{x}_T \in \mathbb{R}^{64{\times}64{\times}3}$ to $\textbf{x}_0 \in \mathbb{R}^{64{\times}64{\times}3}$ which will then be decoded to a synthesized image of size $256{\times}256{\times}3$ using VAE's decoder. In subsequent discussions, we will term $\textbf{x}_T$ as the ``latent code'', and $\textbf{x}_0$ as the ``image'' to avoid confusion between diffusion models' latent space and VAE's latent space.

The text conditions are converted to a sequence of 77 tokens using BERT-tokenizer~\cite{devlin2018bert}, and are then embedded using 32 transformer encoder layers to obtain the $77{\times}640$ text condition embedding. The segmentation masks are downsampled to $32{\times}32$ resolution, and each pixel is expanded to a $1{\times}19$ one-hot vector to encode the 19 classes of facial components. The uni-modal diffusion models are trained with learning rate of $2{\times}10^{-6}$ and batch size of 32 on CelebA-HQ~\cite{karras2018pggan}'s $256{\times}256$ images and corresponding condition annotations.

The \textit{dynamic diffuser} $\textbf{D}_{\theta_m}$ takes the noisy image $\bx_t$, timestep $t$, and the condition $c_m$ as input, and predicts the \textit{influence function} $\textbf{I}_{m,t}$. Since the input noisy image $\bx_t \in \mathbb{R}^{64{\times}64{\times}3}$ and the output \textit{influence function} $\textbf{I}_{m,t} \in \mathbb{R}^{64{\times}64{\times}1}$ has the same spatial resolution, we implement \textit{dynamic diffuser} as a UNet~\cite{ronneberger2015unet}.

The timestep $t$ is injected to the \textit{dynamic diffuser} using Adaptive Layer Normalization (AdaLN)~\cite{ba2016layer}:
\begin{gather}
    h_{out} = (1 + s(t))\textit{LayerNorm}(h_{in}) + b(t),
\end{gather}
where $s(\cdot)$ and $b(\cdot)$ are linear layers that project the timestep $t$ to the scale and bias respectively, and $h_{in}$ and $h_{out}$ are the intermediate activations before and after timestep injection.

The condition $c_m$ is fed into the \textit{dynamic diffuser} via cross-attention \cite{vaswani2017attention} with the intermediate activations $h$:
\begin{gather}
    h_{out} = \textit{CrossAttention}(h_{in}, c_m) = \textit{softmax}(\frac{QK^T}{\sqrt{d}})\cdot V, \\
    Q = W_Q \cdot h_{in}, ~~~ K = W_K \cdot c_m, ~~~ V = W_V \cdot c_m,
\end{gather}
where $W_Q$, $W_K$, and $W_V$ are learnable projection matrices.
We provide an overview of the hyperparameters of \textit{dynamic diffusers} in Table~\ref{tab:dynamic_diffuser_hyperparam}.
\begin{table}[h]
  \centering
    \caption{
        \textbf{Hyperparameters of Dynamic Diffusers}. 
    }       
    \small
    \begin{tabular}{l|c|c}
    \Xhline{1pt}
     & \textbf{Dynamic Diffuser for Text Branch}  & \textbf{Dynamic Diffuser for Mask Branch} \\
     \Xhline{1pt}
    VAE latent space shape & $64{\times}64{\times}3$ & $64{\times}64{\times}3$  \\ 
    Number of parameters & 13.1M & 13.1M \\
    Diffusion steps & 1000 & 1000 \\
    Channels & 32 & 32 \\
    Attention resolutions & 8,4,2 & 8,4,2 \\
    Batch size & 8 samples $\times$ 4 GPUs  & 8 samples $\times$ 4 GPUs \\
    Number of iterations & 187k & 187k \\
    Learning rate & $2{\times}10^{-6}$  & $2{\times}10^{-6}$ \\
    \Xhline{1pt}
  \end{tabular}
  \label{tab:dynamic_diffuser_hyperparam}
\end{table}

\noindent Our \textit{dynamic diffuser} has a much smaller model size than the conditional diffusion model, as shown in Table~\ref{tab:num_param}.

\begin{table}[h]
  \centering
    \caption{
        \textbf{Comparison of Model Size}. 
        A \textit{dynamic diffuser} is much smaller than a uni-modal conditional diffusion model.
    }       
    \small
    \begin{tabular}{l|c}
    \Xhline{1pt}
    \textbf{Model Name} & \textbf{Number of Parameters} \\
    \Xhline{1pt}
    Mask-Driven Pre-trained Diffusion Model & 403.6M \\
    Text-Driven Pre-trained Diffusion Model & 403.6M \\
    Dynamic Diffuser for Mask Branch &  13.1M \\
    Dynamic Diffuser for Text Branch &  13.1M \\
    \Xhline{1pt}
  \end{tabular}
  
  \label{tab:num_param}
\end{table}

\subsection{Collaborative Editing}

In this work, all face editing results, including user study and the qualitative results, are applied on \textbf{\textit{real images}} in the validation split of the CelebA-HQ Dataset.

We use Imagic~\cite{kawar2022imagic} to demonstrate that our \textbf{\textit{Collaborative Diffusion}} framework can be extended from synthesis to editing. Imagic is a text-based image editing method using diffusion models, and involves three steps to complete an edit. Given the input image $\bx_{input}$ and target text $c_{text,target}$, Imagic first optimizes the text condition so that the diffusion model $\bepsilon_{\theta_{text}}$ can reconstruct the input image:
\begin{gather}
    c_{text,opt} = \operatorname*{\arg\!\min}_{c_{text}} \mathbb{E}_{\bepsilon, t} \left\| \bepsilon - \bepsilon_{\theta_{text}}(\bx_{t}, t, c_{text}) \right\|^2, 
\end{gather}
where $c_{text}$ is initialized as $c_{text,target}$ before optimization, and $\bx_{t}$ is constructed using the \textit{diffusion process} via $\bx_{t} = \sqrt{\bar\alpha_t} \bx_{input} + \sqrt{1-\bar\alpha_t}\bepsilon$. 
To further improve the fidelity of the input image, the diffusion model $\bepsilon_{\theta_{text}}$ is then fine-tuned with the optimized condition $c_{text,opt}$ being fixed:
\begin{gather}
    \theta_{text,opt} = \operatorname*{\arg\!\min}_{\theta_{text}} \mathbb{E}_{\bepsilon, t}  \left\| \bepsilon - \bepsilon_{\theta_{text}}(\bx_{t}, t, c_{text, opt}) \right\|^2. 
\end{gather}
Finally, Imagic interpolates between $c_{text,target}$ and $c_{text,opt}$ to obtain the interpolated condition $c_{text,int}$:
\begin{gather}
    c_{text,int} = \alpha \cdot c_{text,target} + (1 - \alpha) \cdot c_{text,opt} .
\end{gather}
The edited image is synthesized using the interpolated text condition $c_{text,int}$ and the fine-tuned diffusion model $\bepsilon_{\theta_{text}}$.

We generalize Imagic to achieve mask-driven editing by optimizing the mask condition embedding $c_{mask, target}$ and fine-tuning the pre-trained mask-driven model $\bepsilon_{\theta_{mask}}$. We then use \textbf{\textit{Collaborative Diffusion}} to integrate any text-driven edit and mask-driven edit on the same input image into a collaborative edit.

\section{Further Explanations on Experimental Details}
\label{sec:experimental}

\subsection{Dataset}

The CelebA-HQ Dataset~\cite{karras2018pggan} consists of 30,000 high-resolution images. 
We use the multi-modal annotations for these images in the CelebAMask-HQ~\cite{lee2020maskgan} Dataset and the CelebA-Dialog Dataset~\cite{jiang2021talktoedit}. The 30,000 images are split into the training set (27,000 images) and validation set (3,000 images). The training of uni-modal diffusion models and the \textit{dynamic diffusers} are conducted on the training set, and all the results reported and shown in this work are using multi-modal conditions from the validation set.

The segmentation masks in the CelebAMask-HQ Dataset has 19 classes including facial components and
accessories: `background', ‘skin’, ‘nose’, ‘left eye’, ‘right eye’, ‘left eyebrow’, ‘right eyebrow’,
‘left ear’, ‘right ear’, ‘mouth’, ‘upper lip’, ‘lower lip’, ‘hair’, ‘hat’, ‘eyeglass’, ‘earring’,
‘necklace’, ‘neck’, and ‘cloth’.

The texts in the CelebA-Dialog Dataset provide fine-grained natural language descriptions of the five attributes: `Bangs', `Eyeglasses', `Beard', `Smiling', `Age'. To avoid conflict between segmentation masks and texts, we trimmed the descriptions regarding `Bangs', `Eyeglasses' and  `Smiling' from the natural language descriptions as they are described by segmentation masks as well.

\subsection{Implementation Details on Comparison Methods}

\vspace{6pt}
\noindent \textbf{TediGAN}~\cite{xia2021tedigan1, xia2021tedigan2}.
\textit{TediGAN} is a StyleGAN-based method for text-driven face generation and manipulation. It can be extended to support other modality's guidance by projecting the conditions into StyleGAN's $\mathcal{W}+$ latent space, and performing style mixing to achieve multi-modal control. 
We use TediGAN~\cite{xia2021tedigan1, xia2021tedigan2}'s official implementation for text-driven and mask-driven generation and editing. For multi-modal driven generation, we mix the style codes of text and mask using TediGAN's style mixing control mechanism. For editing, the style codes are initialized using the inverted $\mathcal{W}+$ codes of the input image, and the remaining steps are the same as generation.

\vspace{6pt}
\noindent \textbf{Composable}~\cite{liu2022composablediffusion}.
Both \textit{Composable} and \textit{Ours} use the same set of pre-trained uni-modal conditional diffusion models described in Section~\ref{sec:implementation_details}.
To accelerate the time-consuming sampling process of diffusion models while maintaining fair comparisons, we use DDIM~\cite{song2020ddim} with 50 steps in all experiments (\ie, quantitative, qualitative, and user study) involving \textit{Composable} or \textit{Ours}.

\section{More Qualitative Results}
\label{sec:qualitative}

We show various qualitative results in Figure~\ref{fig:generation_a}-\ref{fig:influence_b}, which are located at the end of this Supplementary File.

\subsection{Generation and Editing}

\begin{figure*}[h]
  \centering
    \includegraphics[width=0.9\textwidth]{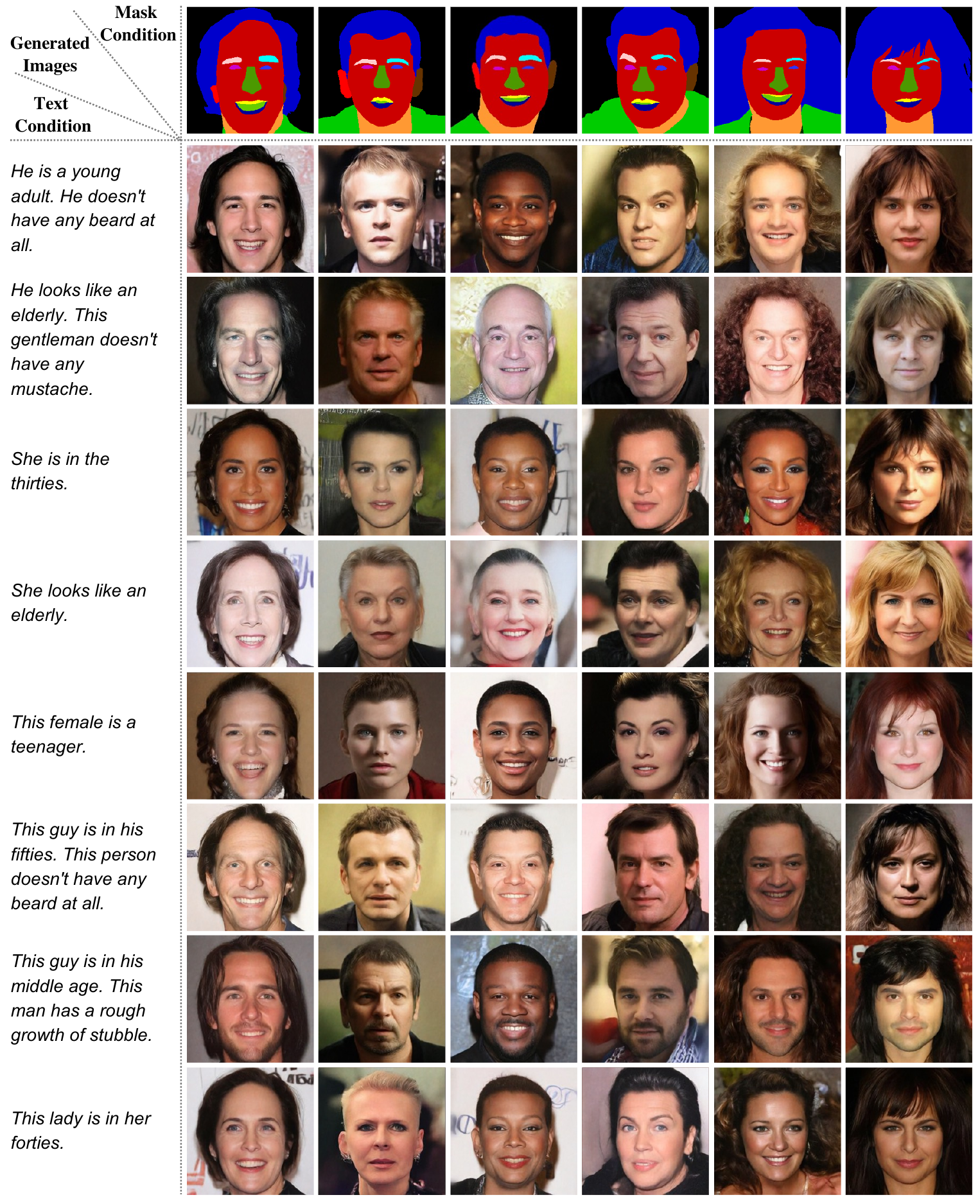}
   \caption{\textbf{More Face Generation Results (A)}. Our method generates realistic images under different combinations of multi-modal conditions, even for relatively rare combinations in the training distribution, such as a man with long hair.}
   \label{fig:generation_a}
    \vspace{-5pt}
\end{figure*}

\begin{figure*}[h]
  \centering
    \includegraphics[width=0.99\textwidth]{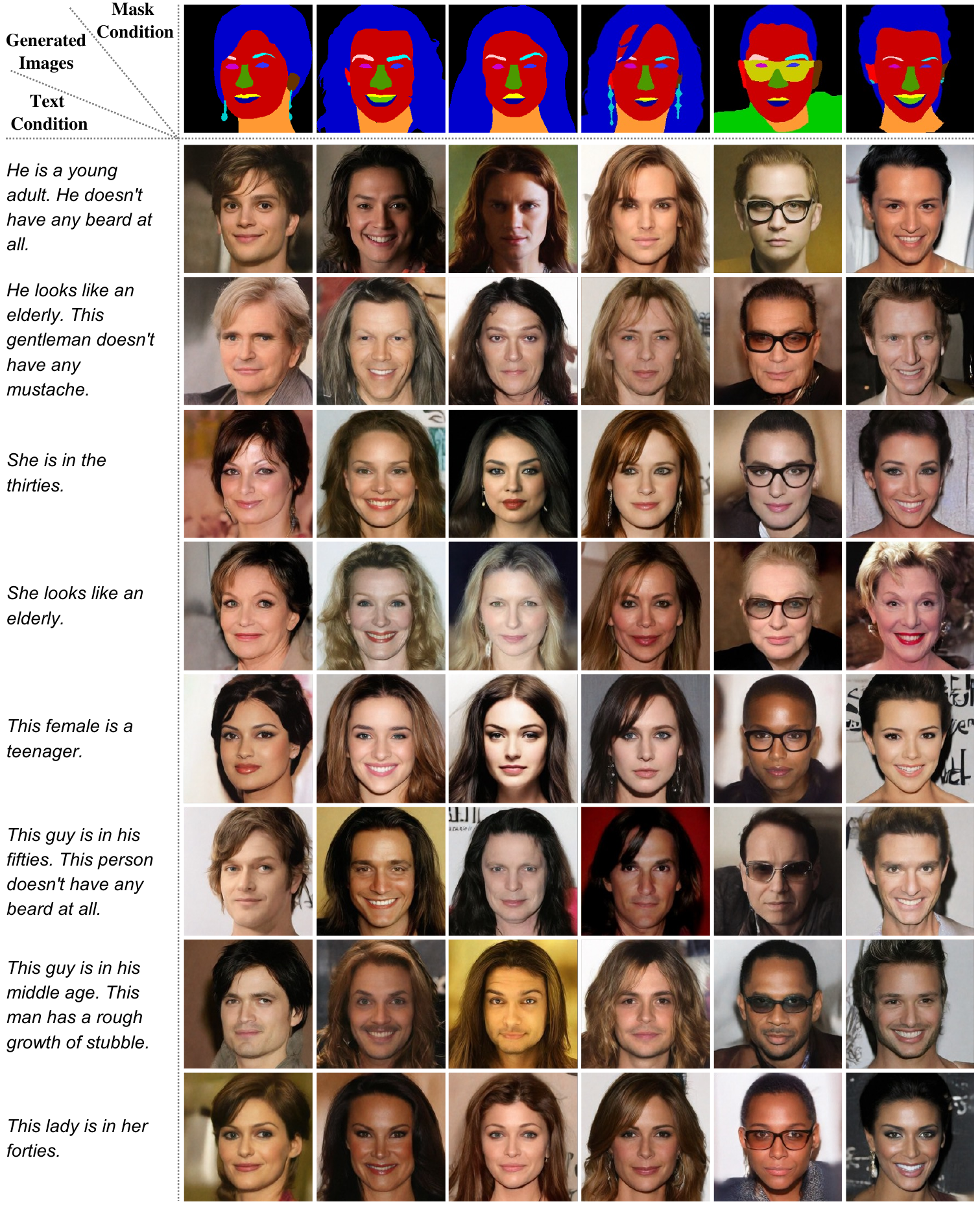}
   \caption{\textbf{More Face Generation Results (B)}. Our method generates realistic images under different combinations of multi-modal conditions, even for relatively rare combinations in the training distribution, such as a man with long hair.}
   \label{fig:generation_b}
    \vspace{-5pt}
\end{figure*}

\begin{figure*}[h]
  \centering
    \includegraphics[width=0.95\textwidth]{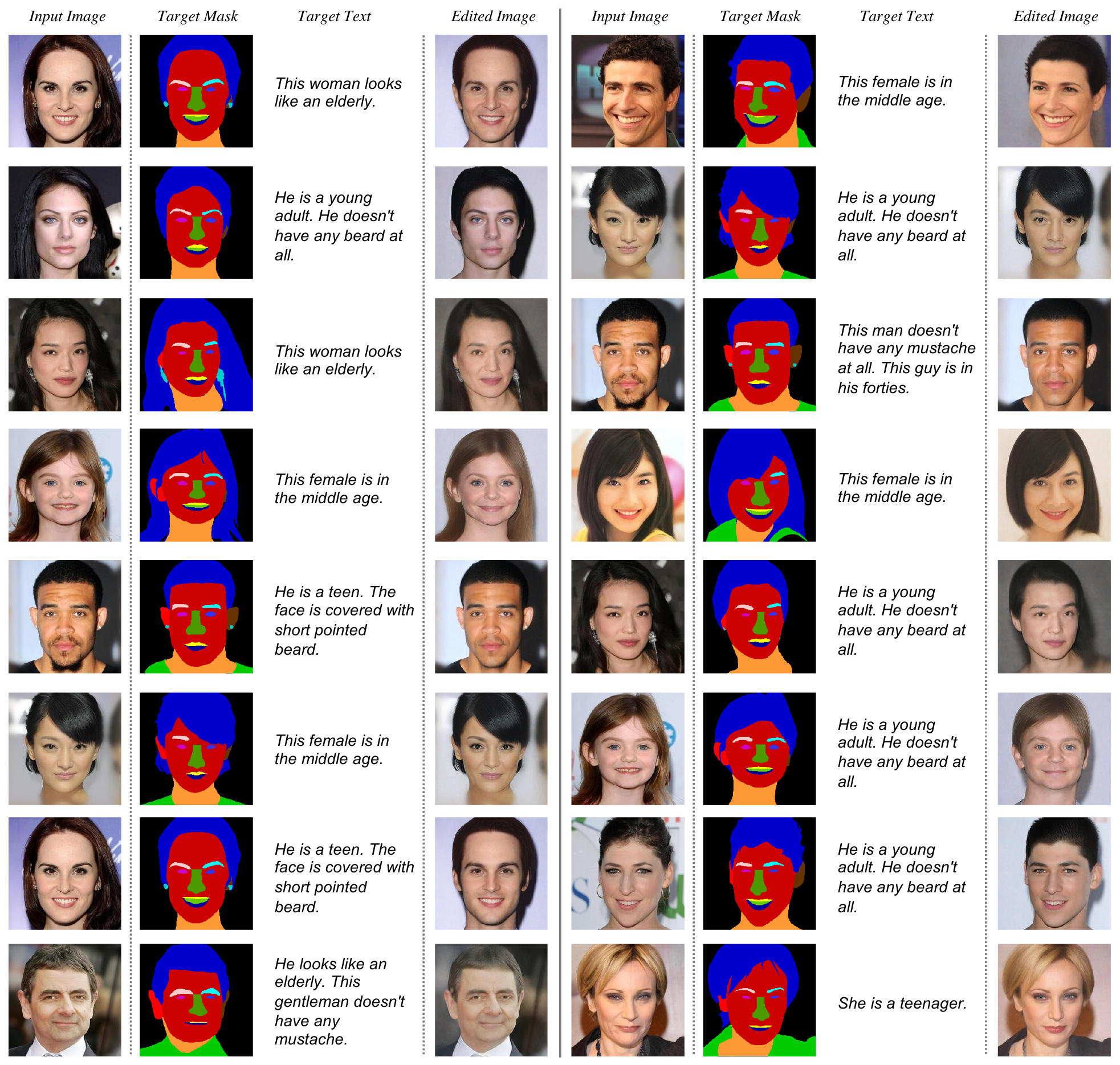}
    \vspace{-5pt}
   \caption{\textbf{Face Editing Results}. Given the input real image and target conditions, we display the edited image using our method.}
   \label{fig:supp_editing}
\end{figure*}

We provide more face generation results in Figure~\ref{fig:generation_a} and Figure~\ref{fig:generation_b}, and face editing results in Figure~\ref{fig:supp_editing}.

\subsection{Visualization of Influence Functions}
\begin{figure}[h]
  \centering
  \includegraphics[width=0.8\linewidth]{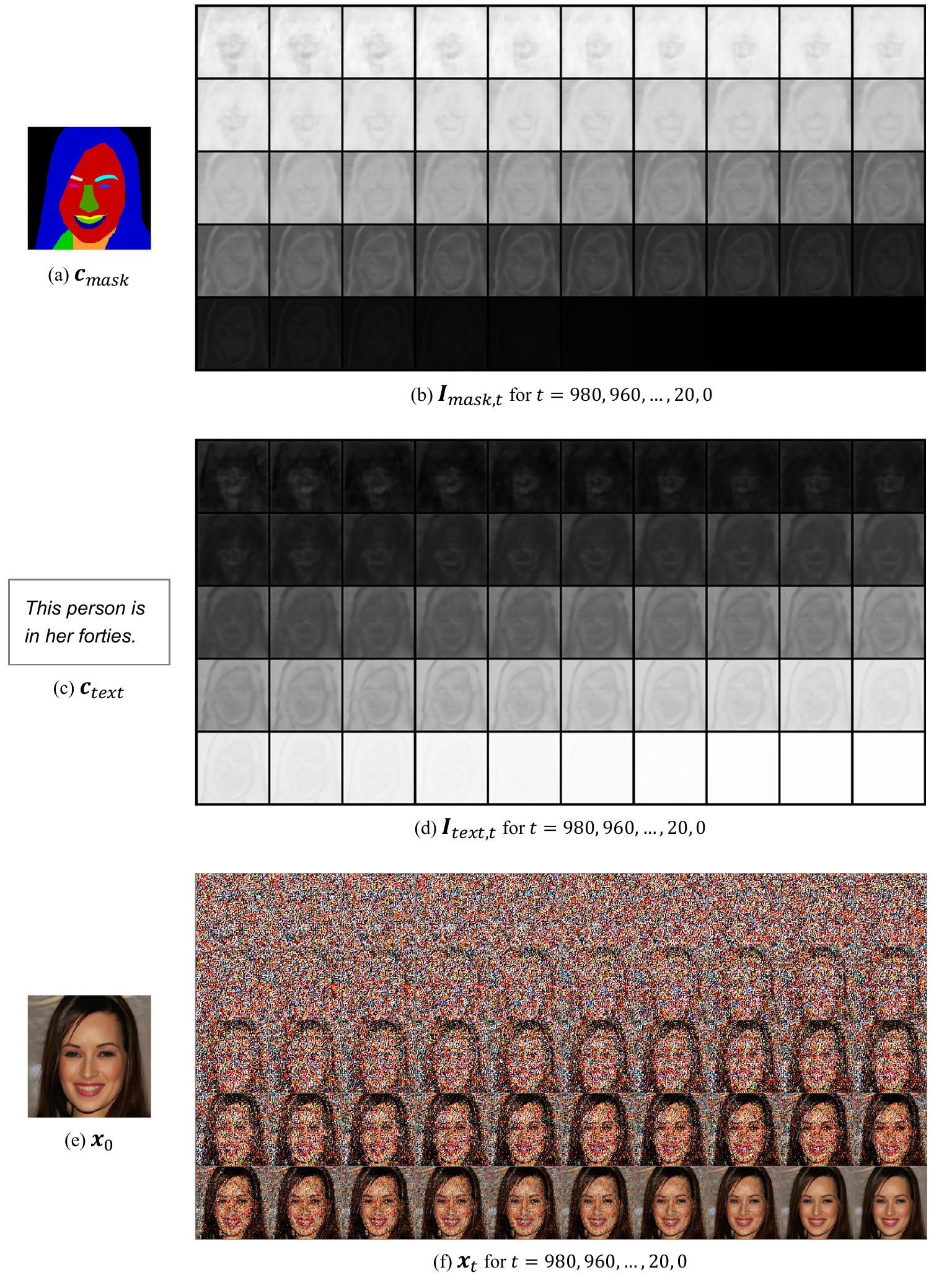}
  \caption{\textbf{Visualization of Influence Functions (A)}. 
  The \textit{influence function} varies spatially at different face regions, and temporally at different diffusion timesteps. The spatial-temporal adaptivity of \textit{influence functions} facilitates effective collaboration.
  }
  \label{fig:influence_a}
  \vspace{-5pt}
\end{figure}

\begin{figure}[h]
  \centering
  \includegraphics[width=0.8\linewidth]{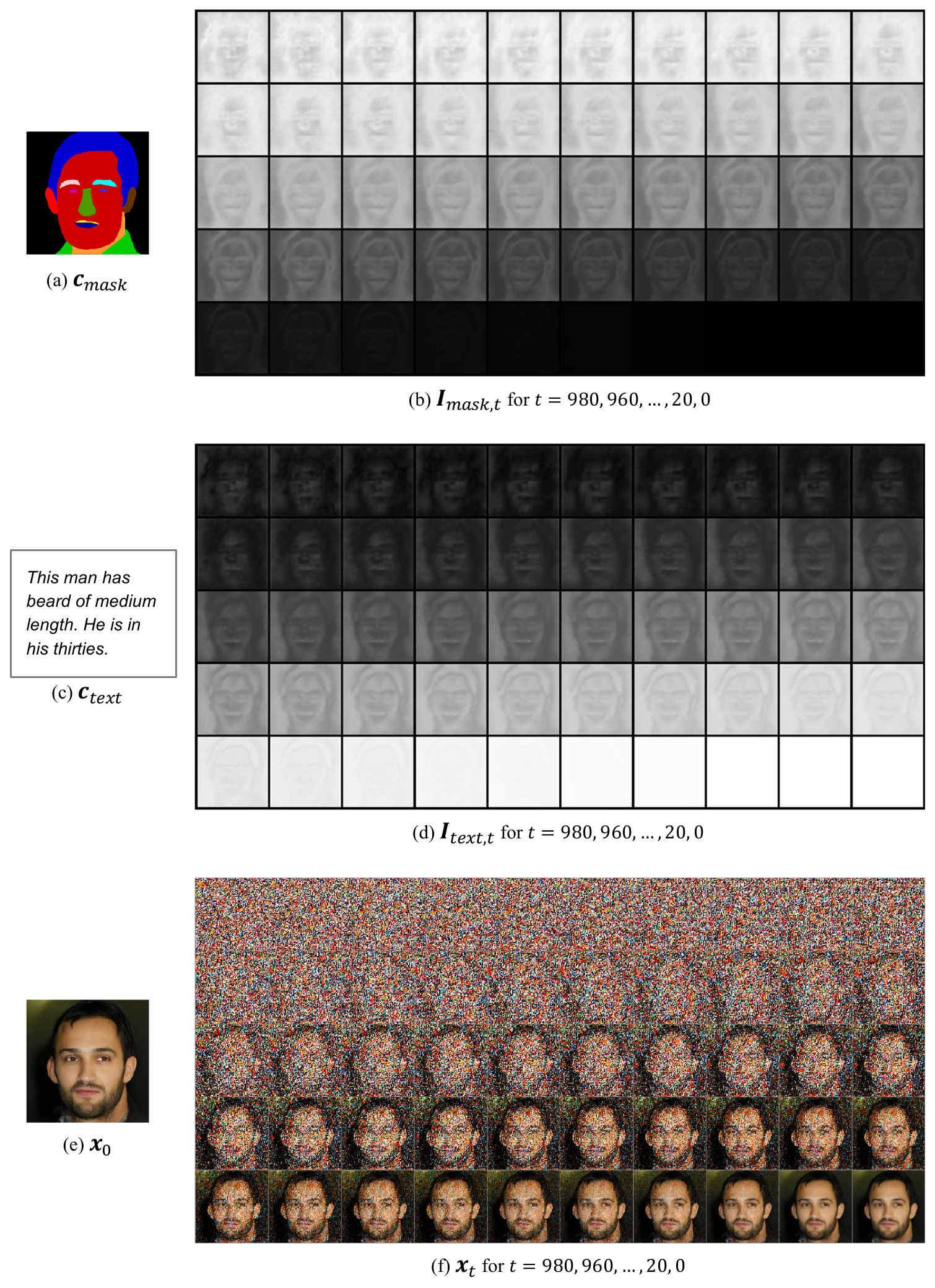}
  \caption{\textbf{Visualization of Influence Functions (B)}. 
  The \textit{influence function} varies spatially at different face regions, and temporally at different diffusion timesteps. The spatial-temporal adaptivity of \textit{influence functions} facilitates effective collaboration.
  }
  \label{fig:influence_b}
  \vspace{-5pt}
\end{figure}

In Figure~\ref{fig:influence_a} and Figure~\ref{fig:influence_b}, we visualize the \textit{influence functions} to show their spatial-temporal variation.
Given the mask condition in Figure~\ref{fig:influence_a}(a), Figure~\ref{fig:influence_a}(b) displays the \textit{influence functions} of the mask-driven collaborator at each DDIM sampling step $t = 980, 960, ..., 20, 0$, from the left to right, top to down. The text branches' \textit{influence functions} are displayed similarly. In Figure~\ref{fig:influence_a}(f), we show the intermediate diffusion results $\bx_t$ for $t = 980, 960, ..., 20, 0$ by decoding them to the image space using the VAE decoder. The final synthesized image is displayed in Figure~\ref{fig:influence_a}(e). Figure~\ref{fig:influence_b} displays the intermediate results using a different set of multi-modal conditions, and is arranged in the same way as Figure~\ref{fig:influence_a}.

\section{Potential Societal Impacts}
\label{sec:impact}

\textit{Collaborative Diffusion} can achieve high-quality real image editing driven by different modalities. However, such capabilities could be applied to maliciously manipulate real human faces. Therefore, we advise users to use \textit{Collaborative Diffusion} only for proper recreational purposes.

The rapid progress in generative models unleashes creativity, but inevitably introduces various societal concerns. First, it becomes easier to create false imagery or maliciously manipulate the data, which could lead to the spread of misinformation. Second, training data might be revealed during the sampling process without explicit consent from data owner~\cite{tinsley2021face}. Third, generative models potentially suffer from the biases present in the training data~\cite{esser2020note}. For \textit{Collaborative Diffusion}, we conducted training on CelebA-HQ~\cite{karras2018pggan}'s faces of various celebrities, which could potentially deviate from the looks of the general population. We hope to see more research to alleviate the risks and biases of generative models, and we advise all to apply generative models with discretion.

\pagebreak

\end{document}